\documentclass[journal]{IEEEtran}

\ifCLASSINFOpdf
   \usepackage[pdftex]{graphicx}
\else
   \usepackage[dvips]{graphicx}
\fi

\usepackage{amsmath}
\usepackage{amssymb}
\usepackage{cite}
\usepackage{mathrsfs}
\usepackage{algorithm}
\usepackage{algorithmic}
\usepackage{subfigure}
\usepackage{xcolor}
\usepackage{bm}
\usepackage{booktabs}
\usepackage{threeparttable}
\usepackage{multirow}
\usepackage{algorithm}
\usepackage{algorithmic}
\begin{document}
\title{Joint Person Objectness and Repulsion for Person Search}
\author{
Hantao Yao, \emph{Member, IEEE}
Changsheng Xu, \emph{Fellow, IEEE}

\IEEEcompsocitemizethanks{

\IEEEcompsocthanksitem Hantao Yao is with National Laboratory of Pattern Recognition, Institute of Automation, Chinese Academy of Sciences, Beijing, 100190, China, Email:hantao.yao@nlpr.ia.ac.cn
\IEEEcompsocthanksitem Changsheng Xu is with National Laboratory of Pattern Recognition, Institute of Automation, Chinese Academy of Sciences, Beijing, 100190, China, and also with the University of the Chinese Academy of Sciences, Beijing 100049, China, Email: csxu@nlpr.ia.ac.cn
}}

\maketitle
\begin{abstract}
Person search targets to search the probe person from the unconstrainted scene images, which can be treated as the combination of person detection and person matching. 
However, the existing methods based on the Detection-Matching framework ignore the person objectness and repulsion (OR) which are both beneficial to reduce the effect of distractor images. 
In this paper, we propose an OR similarity by jointly considering the objectness and repulsion information.
Besides the traditional visual similarity term, the OR similarity also contains an objectness term and a repulsion term.
The objectness term can reduce the similarity of distractor images that not contain a person and boost the performance of person search by improving the ranking of positive samples.
Because the probe person has a different person ID with its \emph{neighbors}, the gallery images having a higher similarity with the \emph{neighbors of probe} should have a lower similarity with the probe person.
Based on this repulsion constraint, the repulsion term is proposed to reduce the similarity of distractor images that are not most similar to the probe person. 
Treating the Faster R-CNN as the person detector, the OR similarity is evaluated on PRW and CUHK-SYSU datasets by the Detection-Matching framework with six description models.
The extensive experiments demonstrate that the proposed OR similarity can effectively reduce the similarity of distractor samples and further boost the performance of person search, \emph{e.g.,} improve the mAP from 92.32\%  to 93.23\% for CUHK-SYSY dataset, and from 50.91\% to 52.30\% for PRW datasets.
\end{abstract}


\IEEEpeerreviewmaketitle

\section{Introduction}
\begin{figure}
\centering
\includegraphics[width=0.9\linewidth]{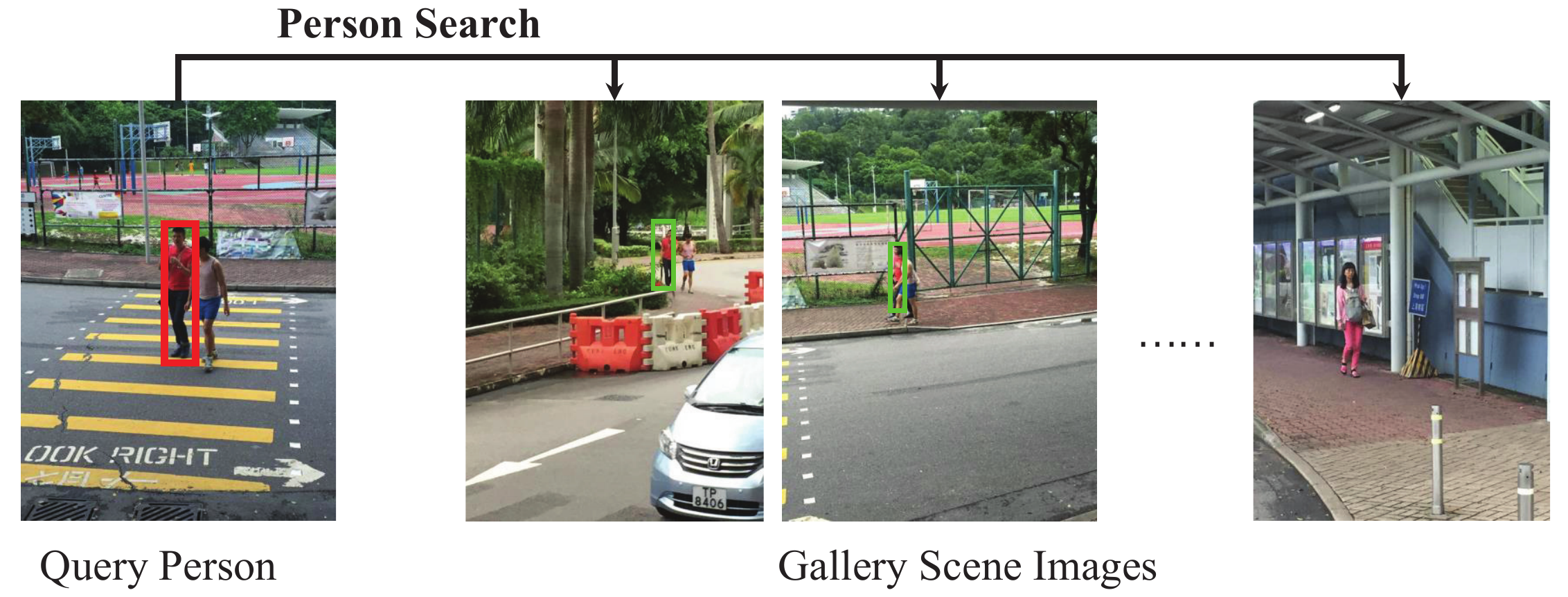}
\caption{Illustration of person search. The red and green bounding boxes denot the query person and the true positive persons, respectively.}
\label{Fig:ps}
\vspace{-0.28cm}
\end{figure}

Given an image along with the probe person, person search targets to search the probe person from an extensive gallery of unconstrainted scene images~\cite{XiaoLWLW17}, as shown in Figure~\ref{Fig:ps}.
An intuitive solution for the person search determines whether the scene image contains a person who has the same ID as the probe person.
Due to the multi-scale and occlusion problems of persons in the scene images, it is almost impossible to directly measure the similarity between the probe person and the unconstrainted scene images.
Therefore, person detection~\footnote{In this paper, the person detection has the same meaning to pedestrian detection.}, which focuses on localizing persons among an image, is a priori component for person search.
Once localizing the candidate person proposals, person search can measure the visual similarity between the probe person and all candidate persons, and find the most matched one.
Therefore, person search can be treated as a combination of person detection and person matching. 

\begin{figure*}
\begin{center}
\includegraphics[width=0.85\linewidth]{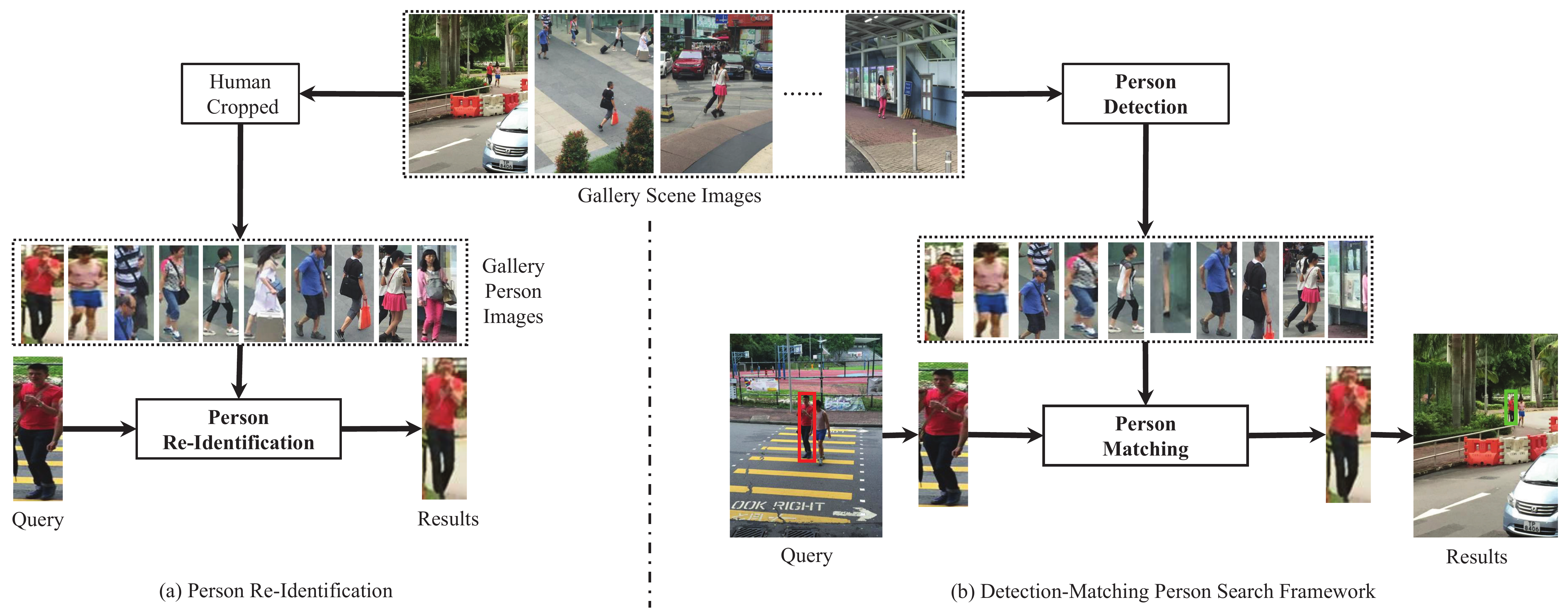}
\end{center}
\caption{The illustration of person re-identification (a) and Detection-Matching person search framework (b).}
\label{Fig:reidandps}
\end{figure*}

Depending on whether using a separate description model for person matching, existing methods can be classified into two categories: Unified person search framework and Detection-Matching framework. 
The unified framework combines person detection and person matching into an end-to-end model~\cite{LiYC18,XiaoLWLW17,LiuFJKZQJY17,DBLP:conf/cvpr/MunjalATG19}. 
However, the subtle difference between different person images makes those methods not to generate a robust and discriminative person description. 
Different from the unified framework, the Detection-Matching framework~\cite{ZhengZSCYT17,LanZG18,ChenZOYT18,DBLP:conf/cvpr/YanZNZXY19} takes person detection and person matching as two independent components.
It firstly employs a person detector to generate the candidate person proposals, and then uses a description model to generate the person representation used for matching, as shown in Figure~\ref{Fig:reidandps}(b).
In current person search methods, person matching is always implemented by person re-identification.
Person re-identification is significantly different from Detection-Matching person search framework because the candidate gallery person images are generated by human cropped in person re-identification, as shown in Figure~\ref{Fig:reidandps}(a). 
Assisted in the success of the detection model, existing methods based on the Detection-Matching framework mainly focus on how to generate a robust and discriminative person description.

\begin{figure}
\includegraphics[width=0.9\linewidth]{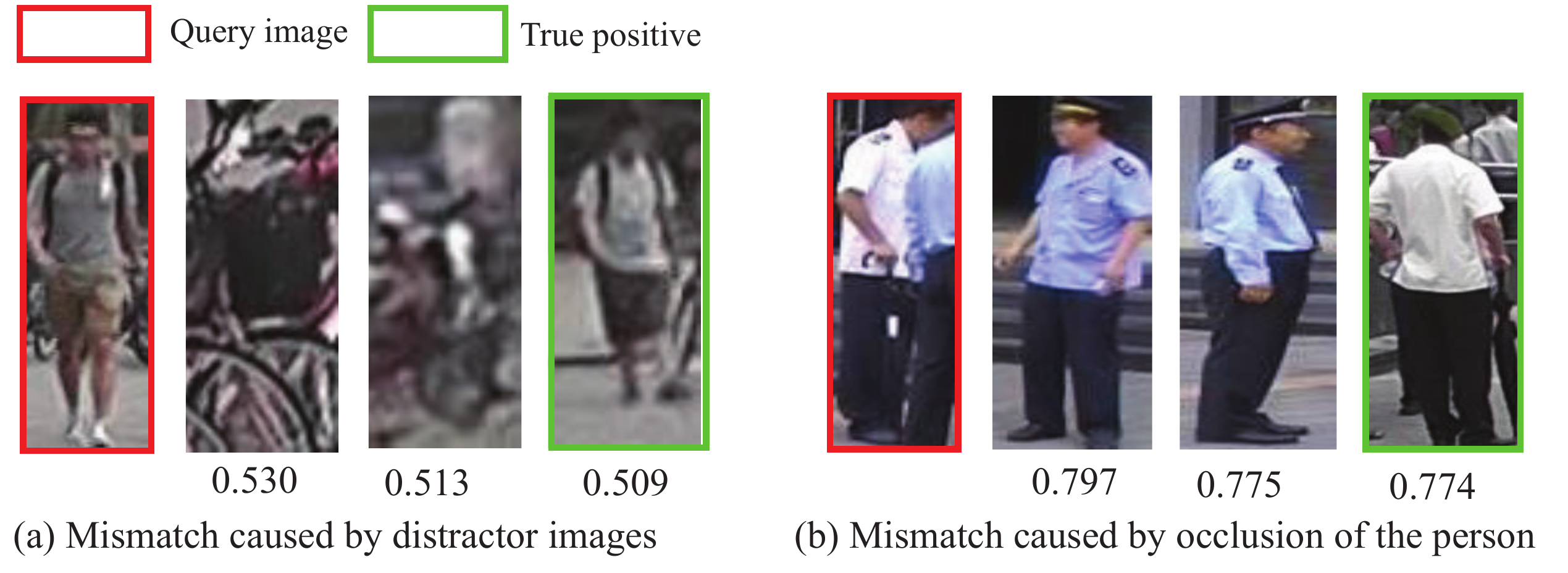}
\caption{The mismatch caused by the distractor images(a), and occlusion of the person(b). }
\label{Fig:samples}
\end{figure}

Although Detection-Matching is a robust framework for the person search, it has the following limitations. 
Firstly, it ignores the objectness of the auto-detected gallery images during person matching.   
The Detection-Matching framework assumes that each gallery image must contain a person, and ignores the objectness information\footnote{The objectness information is the confidence score that the proposal contains a person.}.
Actually, the gallery images are all auto-detected by the person detector,  and may only detect part of the person or even not contain a person.
As a consequence, the distractor images that not contain a person may have a higher similarity than the true positive images.
In conclusion, ignoring the objectness of the auto-detected gallery images brings more false-positive images into the results, as shown in Figure~\ref{Fig:samples}(a).

Secondly, the Detection-Matching framework ignores the repulsion constraint between the probe person and its neighbors.
Given a probe person along with its original frame, we define the~\emph{neighbors} as the auto-detected persons in the original frame.
Based on the prior knowledge that each frame cannot contain two persons with the same ID,  the probe person and its neighbors have a repulsion constraint.
For example, the gallery image having a higher similarity with the neighbors should have a lower similarity with the probe person.
However, the Detection-Matching framework discards the repulsion constraint during person matching, leading to many mismatches caused by the neighbors.
As shown in Figure~\ref{Fig:samples}(b), if its neighbors occlude the probe person, much interference information can be caused by its neighbors, and generate many failure cases. 

To address the above issues, we propose a novel OR similarity by jointly considering the Objectness and Repulsion information of the person. 
Note that the OR similarity is proposed based on the Detection-Matching person search framework.
Besides the traditional visual similarity term, the OR similarity also contains the objectness term and the repulsion term.  
An intuitive illustration is shown in Figure~\ref{Fig:framework}. 
For the Detection-Matching framework, we first train the Faster R-CNN~\cite{RenHG017} as the detector to generate the gallery images. 
Based on the detection score for each gallery image, the objectness term represents the probability that the gallery image contains a person. 
Using the objectness term can boost the performance of person search by reducing the similarity of distractor images and improving the ranking score for positive samples.
With the constraint that the probe person has a different person ID with its neighbors, the repulsion term is proposed to reduce the similarity of gallery images that are not most similar to the probe person. 
For each gallery image, the repulsion term is implemented based on the gap between the similarity of the probe and the nearest neighbors, where the nearest neighbor is the neighbor which has the highest similarity score with the gallery image.

We finally evaluate the proposed OR similarity with the Detection-Matching framework, which treats the Faster R-CNN as detection model and employs six description models for person matching, \emph{i.e.,} ResNet50\cite{HeZRS15}, DenseNet121\cite{HuangLMW17}, Squeeze-and-Excitation Networks (SE-Net)\cite{HuSS18}, Multi-Level Factorisation Net (MLFN)\cite{ChangHX18}, Part-based Convolutional Baseline (PCB)\cite{SunZYTW18}, Multiple Granularities Networks(MGN)\cite{WangYCLZ18}. 
The extensive experiments demonstrate that the OR similarity is compatible with existing Detection-Matching person search methods, and can further boost their performance. 
The contributions of this paper are summarized as follows:
\begin{enumerate} 
\item We show that considering the objectness of each gallery image can reduce the effect of distractor images for the person search. 
\item We demonstrate that using the repulsion between probe person and its neighbors can decrease the similarity of the distractor images that are not most similar to the probe person.
\item By jointing the person objectness and repulsion information, we propose an OR similarity that is compatible with existing methods and can further boost their performance.
\end{enumerate}

\section{Related Work}
In this section, we review the related work for person search from the following three aspects: person search, person detection, and person re-identification.

\paragraph{Person Search}
Recently, the person search, which targets to search the probe person from a large scale of gallery whole scene images, has drawn much attention. As the person search can be regarded as the combination of person detection and person re-identification, the existing methods can be classified into two classes: 1) end-to-end person search framework~\cite{LiYC18,XiaoLWLW17,LiuFJKZQJY17,DBLP:conf/cvpr/MunjalATG19}; 2) Detection-Matching framework~\cite{LanZG18,ChenZOYT18,DBLP:conf/cvpr/YanZNZXY19}. 

The end-to-end framework aims to propose a unified model to learn the person detection and description jointly. 
For example, based on the Faster R-CNN, Xiao~\emph{et al.}~\cite{XiaoLWLW17} propose an online-instance matching loss to infer the person detection and description model together.
Li~\emph{etal.}~\cite{LiYC18} further propose a Pedestrian Space Transformer to improve the accuracy of person detection, and further boost the performance for person search. 
Different from the above two methods, Liu~\emph{et al.}~\cite{LiuFJKZQJY17} propose a Neural Person Search Machines that can recursively achieve the person detection and searching simultaneously.
As the person image among person search contains subtle difference, the major disadvantage of using a unified model is that it cannot generate a robust and discriminative representation to describe the person. 
Therefore, the Detection-Matching framework treats the person detection and person description as two independent models, and boost the person search performance by improving the representation ability. 
For example, Zheng~\emph{et al.}~\cite{ZhengZSCYT17} employ a detection model for person detection, and a re-identification model for generating a robust person description. 
As the auto-detected person images contain many noisy backgrounds information, Chen~\emph{et al.}\cite{ChenZOYT18} propose a mask-guided two-stream CNN model to generate a robust person description with the help of person foreground mask. Further, the auto-detected person proposals vary in scale, Lan~\emph{et al.}~\cite{LanZG18} propose a Cross-Level Semantic Alignment to improve the descriptive ability, and then employ a multi-scale matching to overcome the scale problem. Therefore,  state-of-the-art person re-identification models can be used to generate person representation. 

\paragraph{Person Detection}
Person detection plays a crucial role in the Detection-Matching framework for person search. With the development of the Convolutional Neural Network, the recent person detection methods are all based on deep learning~\cite{ZhangWBLL18,NohLKK18,SongSXSP18,LiLSXFY18,OuyangW13,ZhouY18,DBLP:journals/tmm/WangCLWZ18}. Based on the previous Deformable Part Models (DPM)~\cite{FelzenszwalbGMR10}, Ouyang~\emph{et al.}~\cite{OuyangW13} propose JointDeep model which employs the deep learning to generate a robust description for the person parts. More recently, as the fantastic performance achieved by the RCNN framework~\cite{GirshickDDM14,Girshick15,RenHG017} in object detection, most of the person detection methods~\cite{ZhangWBLL18,NohLKK18,SongSXSP18,LiLSXFY18,ZhouY18} are all proposed based on the RCNN framework.  
Furthermore, the previous work~\cite{LanZG18,ChenZOYT18} has been shown that the Faster R-CNN is a robust framework for the existing person search datasets, we then employ the Faster R-CNN as the person detector in this paper. 
Recently, the DLR~\cite{han2019re} proposes an ROI transform layer to connect the person detection and person description. With the ROI transform layer, the person description loss can be used to supervise the person detector. Finally, a more robust person detector is applied in DLR for person detection to boost the performance.

\paragraph{Person Re-Identification} The person search, which can be regarded as an extension of the person re-identification by considering the person detection, is the most related to the person re-identification. As the success of person detection, one kind of methods~\cite{ChenZOYT18,LanZG18} for person search are all focus on how to generate a discriminative and robust person description. For the person re-identification, methods~\cite{YiLLL14,LiZXW14,AhmedJM15,VariorSLXW16,ChenZW16,DingLWC15,ChengGZWZ16,WangZLZZ16,WangYCLZ18,DBLP:journals/tmm/ZhouWSHGZ18} focus on employing the siamese or triplet metric learning to improve the discriminative of the generated features. The other methods~\cite{LiC0H17,SunZYTW18,WangYCLZ18,YaoZHZX019,ZhaoLZW17,ZhengHLY19,ZhaoTSSYYWT17,SuLZX0T17,DBLP:journals/tmm/WeiZY0019,ZhouWMLGZ19} focus on enhancing the discriminative by considering the local person part. Targeting to describe the discriminative region of the person image, some methods~\cite{LiuFQJY17,LiZG18,LiuZTSSYYW17} are proposed based on the attention mechanism.

Different from the related person search methods, our work focuses on the measurement between probe image and gallery images, and is independent of the detection model and description model. 
Therefore, we treat the Faster R-CNN as detector, and employ several re-identification models~\cite{ChangHX18,SunZYTW18,WangYCLZ18} to evaluate the effect of our method.

\section{Analysis about existing method}
Before describing the proposed OR Similarity, we first analyze the existing studies on person search from the following two aspects: 1) what is the effect of objectness for person search? 2) what is the effect of repulsion constraint? The following analyses are performed on the PRW dataset~\cite{ZhengZSCYT17,LiuZTSSYYW17,LiZG18} with Faster R-CNN model~\cite{RenHG017}.

\begin{figure}
\begin{center}
\includegraphics[width=1\linewidth]{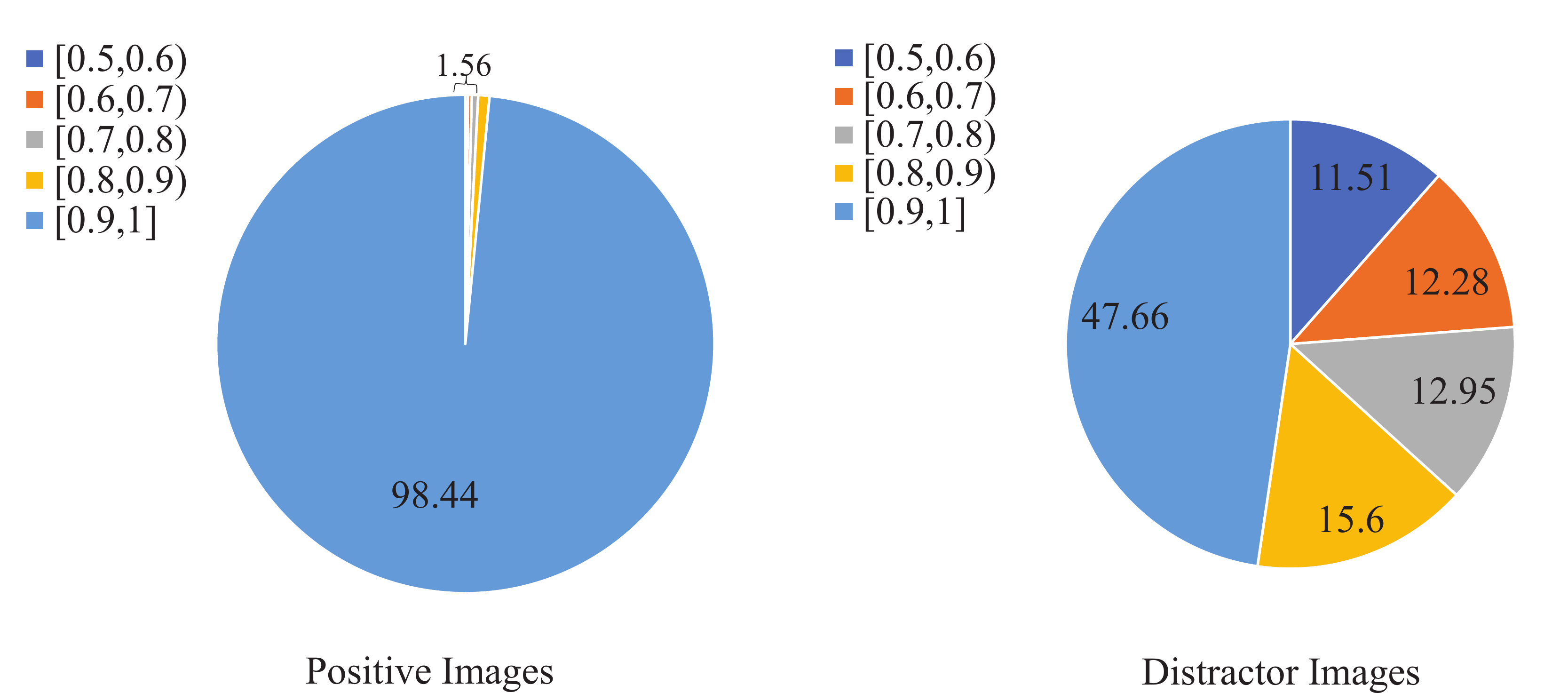}
\end{center}
\caption{The distribution statistics of detection scores for gallery images on PRW dataset. The \emph{Positive} denotes the gallery images which belong to one of the probe persons. The \emph{Distractor} represents the images that do not belong to any probe persons. The detection scores are derived from the person detector.}
\label{Fig:objectness_analysis}
\end{figure}

\subsection{Analysis on Objectness}
Inspired by person re-identification,  the existing Detection-Matching person search framework ignores the objectness of the auto-detected gallery images during person matching. 
The gallery images are generated by manually cropped or carefully filtered from auto-detected bounding boxes for person re-identification. 
Consequently, those gallery images have a high probability of containing a person, and we can ignore the objectness of the gallery image while computing the similarity between probe and gallery images. 
Different from person re-identification, the gallery images of person search are all generated by a person detector. For object detection, the detector treats the proposal with a detection score larger than or equal to 0.5 as the correct detection. 
Therefore, the detection scores of the auto-detected gallery images arrange from 0.5 to 1.0. The higher the detection score, the higher the probability of containing a person, and vice versa. However, in the existing person search methods, the auto-detected proposals are all assumed containing the person with the confidence of 1.0, which is an unreasonable assumption.

\begin{figure}
\begin{center}
\includegraphics[width=1\linewidth]{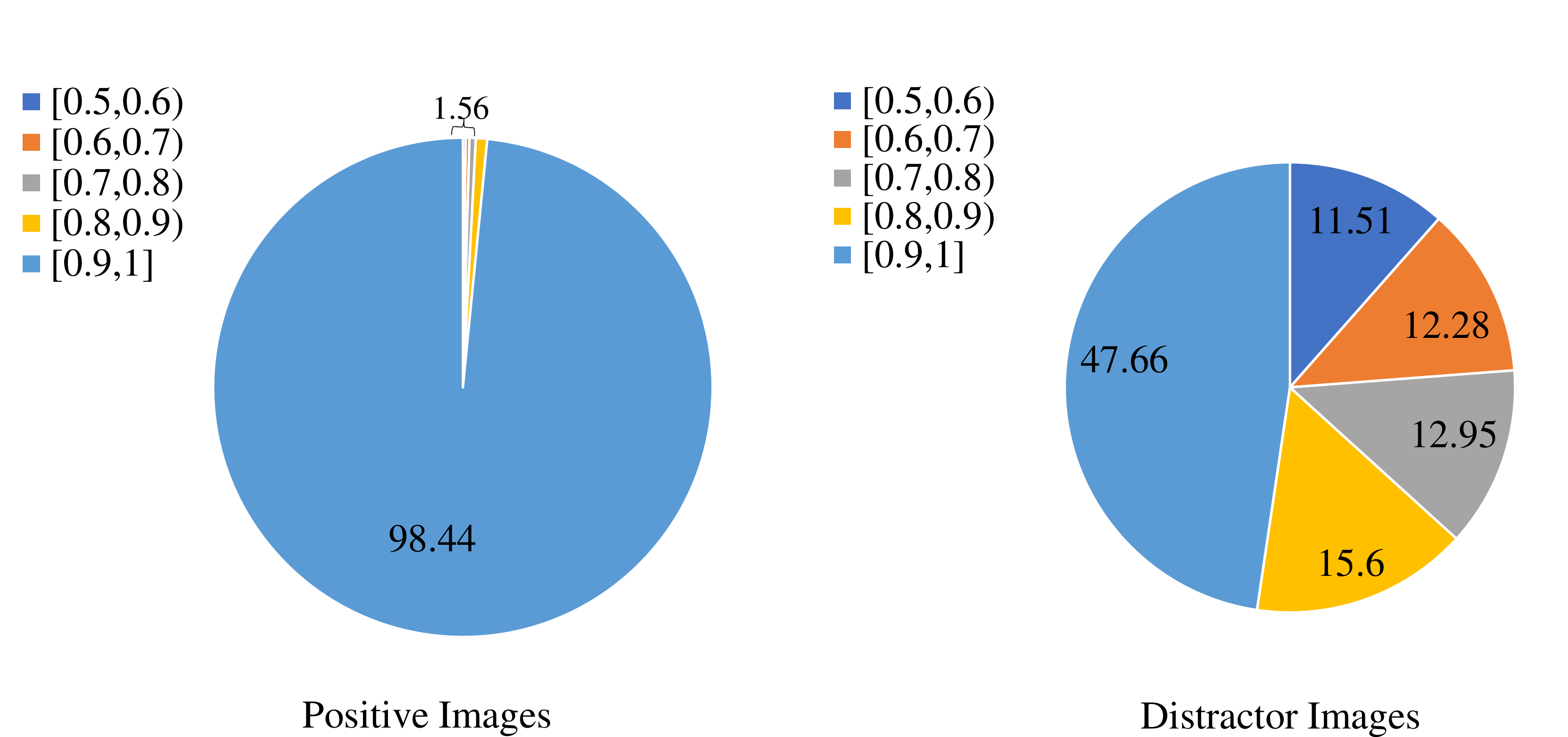}
\end{center}
\caption{The distribution statistics of detection scores for gallery images on CUHK-SYSU dataset. The \emph{Positive} denotes the gallery images which belong to one of the probe persons. The \emph{Distractor} represents the images that do not belong to any probe persons. The detection scores are derived from the person detector.}
\label{Fig:objectness_cuhk}
\end{figure}

\begin{figure}
\begin{center}
\includegraphics[width=0.8\linewidth]{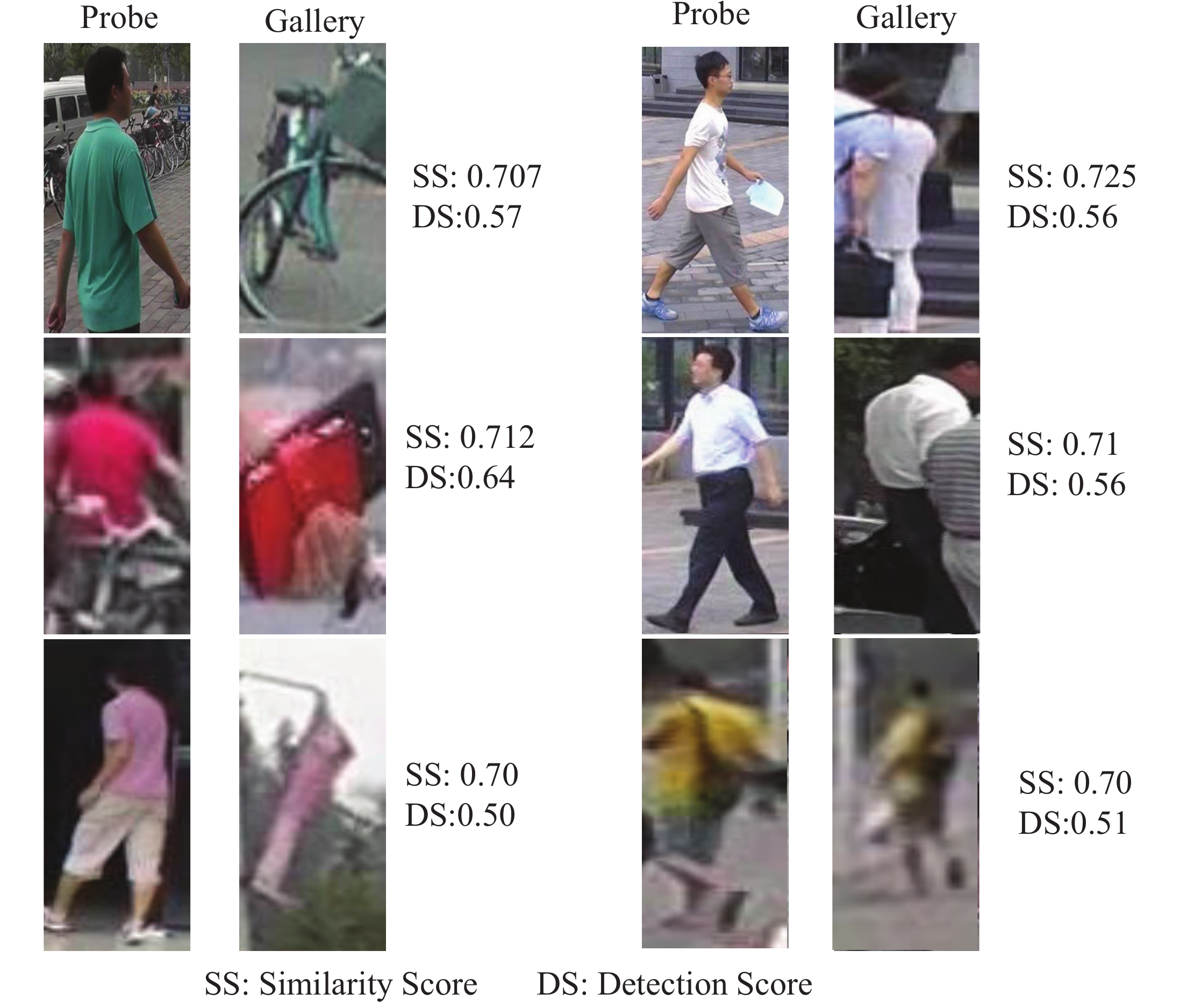}
\end{center}
\caption{The distractor images with a higher similar scores. Note that the averaged similar score between query images and the positive gallery is 0.7.}
\label{Fig:objectness_show}
\end{figure}

Target to analyze the effect of objectness, we statistic the distribution of detection score and show the mismatches in the following. 
We first train a person detector based on the Faster R-CNN on the PRW dataset, and then perform the trained detector to generate the auto-detected person proposals. 
For the PRW dataset, there are 52,942 auto-detected person proposals. 
Based on whether the proposals belong to one of the probe persons, we divided all the detected proposals into two groups: 15,962 positive gallery images, and 36,980 distractor images. 
The positive gallery images are the proposals that have an overlap larger than 0.5 with any ground-truth probe persons, and the distractors are the images that do not belong to any probe persons. 
As shown in Figure~\ref{Fig:objectness_analysis}, the positive gallery images have a high objectness, \emph{e.g.,} more than 98\% of the images have the scores of more than 0.9.
The reason is that the positive gallery images are all carefully annotated with the human.
 Different from the positive gallery images, the distractor images can be classified into three groups: the unlabeled pedestrians, the part of pedestrians, and noisy images. 
 Among all those three types of images, only the unlabeled pedestrians, which contain the pedestrians but not belonging to any probe pedestrians, would have a higher detection score by person detector. 
 The other two types of images both would have a lower detection score due to they both treated as the negative samples while inferring the person detector. 
Therefore, there are half of the distractor images (the unlabeled pedestrian) having higher detection scores, and the other half of distractor images (true negative samples) have lower detection scores.
 As shown in Figure~\ref{Fig:objectness_analysis},  the distractor images contain more than 50\% of images whose detection scores are lower than 0.9. 
 Unfortunately, those images with lower detection scores will produce many mismatches. 
 As shown in  Figure~\ref{Fig:objectness_show}, some samples have low detection scores but a high similarity. Note that the averaged similar score between all query images and its ground truth is 0.7 for the PRW dataset. Therefore, reducing the effect of distractor images by considering the objectness of the proposal can improve the robustness of the person search.

Besides the PRW dataset, we further provide the analysis on the other CUHK-SYSU dataset and illustrate the related results in Figure~\ref{Fig:objectness_cuhk}. From Figure~\ref{Fig:objectness_cuhk}, we can observe that almost all positive images have a higher detection score, e.g., 99.59\% samples have the detection scores than 0.9. For the distractor images, we found that the detection score of more than 60\% samples are also high. 
The reason is that a large number of the unlabeled pedestrians existed in the CUHK-SYSU dataset. Besides the unlabeled pedestrians, the other distractor images all have lower detection scores.

\subsection{Analysis on Repulsion Constraint}
Person matching is performed by computing the similarity between the probe and all gallery images, which is similar to  person re-identification. Compared with person re-identification, the advantage of person search is that the original frame of probe image is available. Given the probe image along with its original frame, we define the other person proposals within the frame as its \emph{neighbors}.
By taking the probe and its neighbors into consideration, we can treat the person matching among person search as a multi-query problem.  As the persons that appear in the same frame must have different IDs, those multiple query images are mutually exclusive. 
For each gallery image, if it has a higher similarity with the neighbors, it should have a lower similarity with the probe person. 
That is to say, the true positive gallery images should have a higher similarity with the probe image than its neighbors. 
Based on the above assumption, we analyze the person search results generated by the existing methods that ignore the repulsion constraint. 
The analysis is performed by taking the state-of-the-art person re-identification model MGN~\cite{WangYCLZ18} as a feature extractor. 
For the PRW with the gallery size of 200, there are 28,105 pairs satisfying the above condition, while the rest 49,017 do not. Therefore, there are most of the true positive gallery images which have a higher similarity with the neighbors.

By further visualizing the results of the person search, we find that the occluded probe person always has a worse search performance. The reason is that the occluded probe person contains the interference information from its neighbors, and has a high similarity to the neighbor that occludes it. An intuitive description is shown in Figure~\ref{Fig:repul}. Once taking the repulsion between the probe image and its neighbors into consideration, it can reduce mismatches caused by the neighbors.

\section{Similarity based on Objectness and Repulsion}
In this section, we introduce the OR similarity that considers the person Objectness and Repulsion. 
We first give the formulation of OR similarity, and then provide a detailed description for each term.

\begin{figure}
\includegraphics[width=1\linewidth]{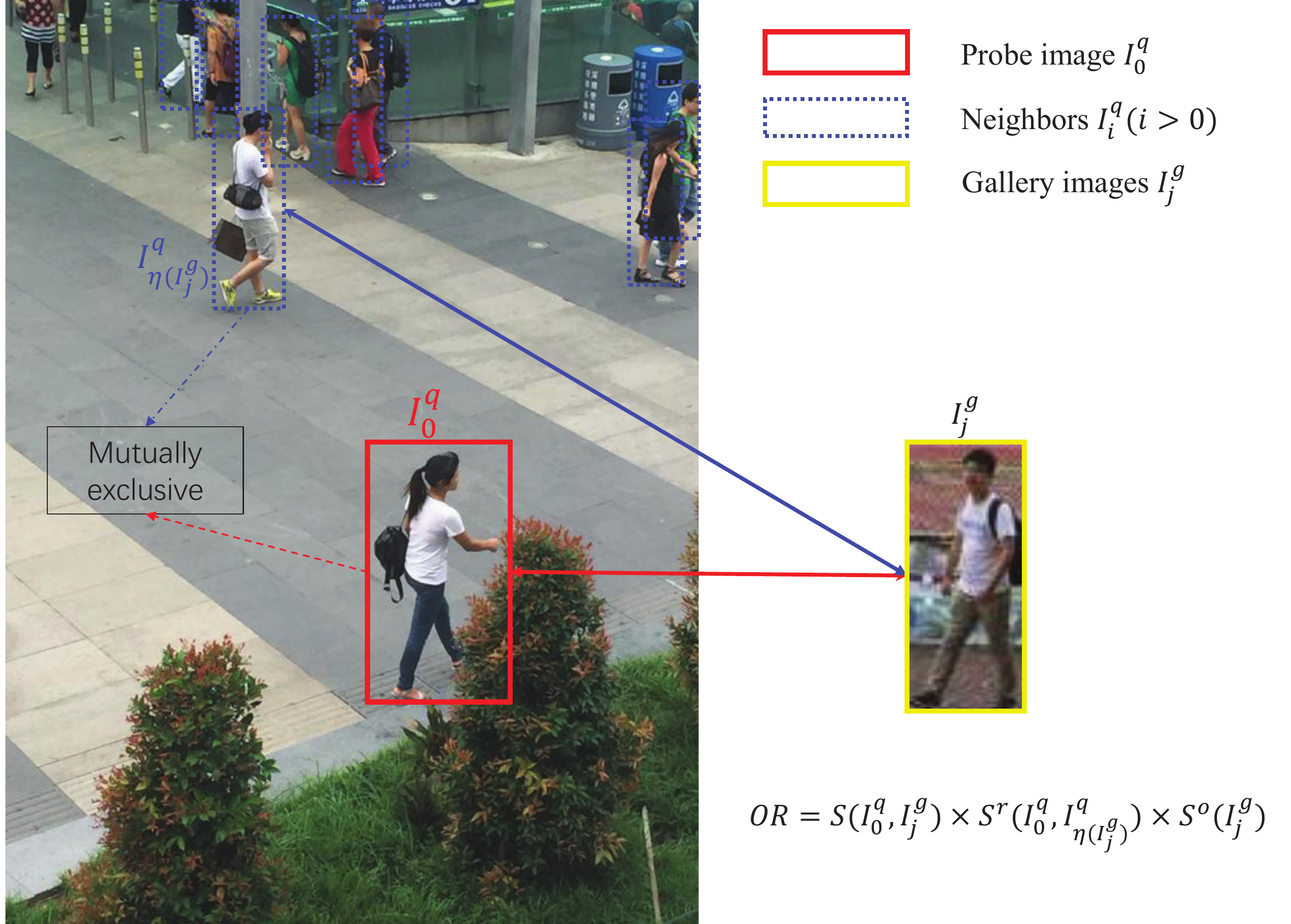}
\caption{The description of OR similarity. $S$ is the visual similarity between two features. $S^{r}$ is the repulsion term, representing the repulsion between the probe image and its neighbors. $S^{o}$ denotes the objectness of gallery image.}
\label{Fig:framework}
\end{figure}

\paragraph{Formulation}
Since the calculation of the similarity between the probe image and each gallery image is independent of person search, we consider only one probe image and one gallery frame in the following for simplicity. 
We define a set of probe image and its neighbor as $\boldsymbol{I}^{q}=\{I^{q}_{0},I^{q}_{1},...,I^{q}_{n_{q}}\}$, where $I^{q}_{0}$ is the probe image, and others are its neighbors. $n_{q}$ is the number of the neighbor images. The gallery images are defined as $\boldsymbol{I}^{g}=\{I^{g}_{1},...,I^{g}_{n_{g}}\}$, where $n_{g}$ is the number of the gallery images. 

The goal of OR term generates a reasonable and robust similarity between $I^{q}_{0}$ and $\boldsymbol{I}^{g}$ with the help of $\boldsymbol{I}^{q}$. By taking the objectness and repulsion constraints into consideration, the OR similarity is defined as Eq.~\eqref{Eq:or}, and an illustration of OR similarity is shown in Figure~\ref{Fig:framework},
\begin{equation}
\label{Eq:or}
S^{or}(I^{q}_{0},I^{g}_{j})=S(I^{q}_{0},I^{g}_{j})\times S^{r}(\boldsymbol{I}^{q},I^{g}_{j})\times S^{o}(I^{g}_{j}),
\end{equation}
where $S$ is the traditional \emph{visual term} which denotes the similarity between the probe and gallery features. $S^{r}$ is the \emph{repulsion term} which denotes the repulsion between the probe image and its neighbors. The $S^{o}$ is the \emph{objectness term} which represents the probability that the gallery image contains a person. $S^{r}\in [0,1]$ and $S^{o}\in[0,1]$ are used to reduce the similarity of negative  gallery and distractor images. Therefore, $S^{r}$ and $S^{o}$ need to have a high score for the positive gallery images, and low score for others. In the following, we give a detailed description of each term.

\paragraph{Visual Term} The visual term $S$ is used to measure the visual similarity between probe image and gallery images. Given the query image $I^{q}_{0}$ and the gallery image $I^{g}_{j} (j\in[1,n_{g}]$), we employ the person description model $f$ to generate the feature for each image, and denote the generated features as $f^{q}_{0}$ and $f^{g}_{i}$. 
To demonstate the effectiveness of the proposed method, we employ six type of models as feature extractor $f$ in this paper, \emph{i.e.,} ResNet50~\cite{HeZRS15}, DenseNet121~\cite{HuangLMW17},Squeeze-and-Excitation Networks~\cite{HuSS18}, Multi-Level Factorisation Net(MLFN)~\cite{ChangHX18},Part-based Convolutional Baseline (PCB)~\cite{SunZYTW18}, Multiple Granularities Networks(MGN)~\cite{WangYCLZ18}. Once obaining the $f^{q}_{0}$ and $f^{g}_{j}$, we define the Euclidean distance between the $L_{2}$ normalized features as the visual term:
\begin{equation}
\label{Eq:S}
S(I^{q}_{0},I^{g}_{j})=1-\frac{1}{2}||f^{q}_{0}-f^{g}_{j}||^{2},
\end{equation}
where $f^{q}_{0}$ and $f^{g}_{j}$ are the description for $I^{q}_{0}$ and $I^{g}_{j}$, respectively.
The more similar images, the higher $S(I^{q}_{0},I^{g}_{j})$ score. $S(I^{q}_{0},I^{g}_{j})$ is also the baseline in our work.

\paragraph{Objectness Term} During person matching, ignoring the objectness of gallery images can damage the performance for person search, especially for the distractor images which do not contain a person. 
Therefore, we employ the objectness term as an additional constraint to reduce the similarity for distractor images. The objectness term should have the following characteristics: 1) it can represent the probability that the image contains a person. 2) its value should be in the range [0,1], the low value corresponding to the distractor images. Therefore, we define the objectness term as follows:
\begin{equation}
S^{o}(j)=e^{S^{d}(j)-1},
\label{eq:ot}
\end{equation}
where $S^{d}(j)$ is the detection score for proposal $I^{g}_{j}$. Inspired by the object detection model, the higher $S^{d}(j)$, the higher the probability that the proposal contains a person. By using $S^{o}(j)$, we can effectively reduce the similar score for the images not belonging to the person category. The advantage of the objectness term is that it can reduce the similar score for distractor images and does not affect the similarity of true positive samples. 
For true positive samples, $S^{d}$ is close to 1 which makes $S^{o}$ close to 1. Therefore, $S^{o}$ would not adjust the similarity for true positive samples. Different from higher $S^{d}$ for true positive samples, the distractor images have a lower $S^{d}$ that produces a lower $S^{o}$.  From the Eq.\eqref{Eq:or}, we can observe that using the lower $S^{o}$ can effectively reduce the similarity.  Figure~\ref{Fig:effect} represents the change of similarity in positive and distractor samples from the objectiveness term.

\paragraph{Repulsion Term} As described above, the probe image $I^{q}_{0}$ has a different person ID with its neighbors $I^{q}_{i} (i\in[1,n_{q}]$). 
Therefore, the gallery image that has a higher similarity with the neighbor should have a lower similarity with the probe image $I^{q}_{0}$. 
To achieve this goal, we propose a repulsion term to reduce the similarity for each gallery image by considering its similarity with the neighbors $\boldsymbol{I}^{q}$.

\begin{figure}
\includegraphics[width=1\linewidth]{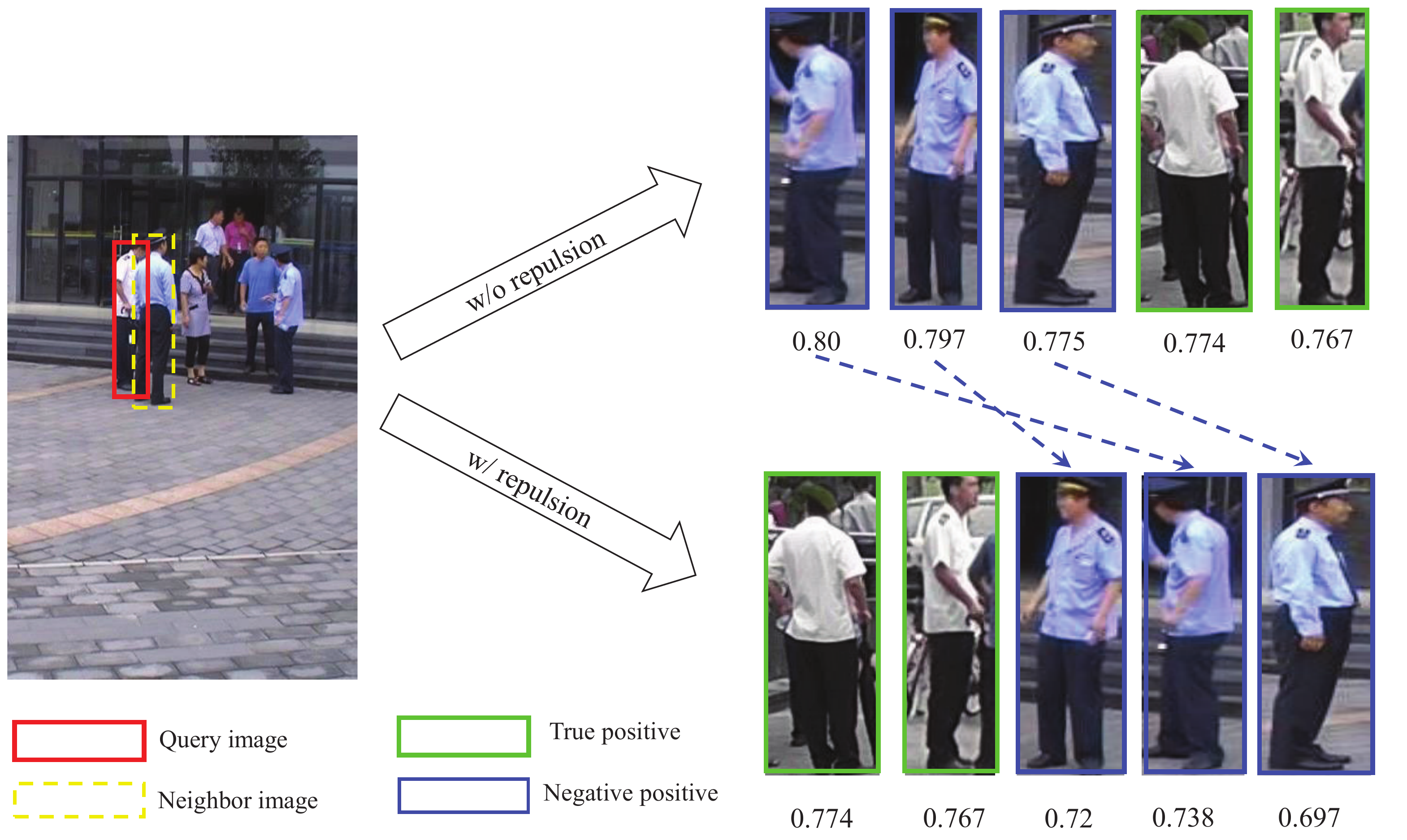}
\caption{The effect of repulsion term. The repulsion term can effectively reduce the similarity for the negative images.}
\label{Fig:repul}
\end{figure}

Given a gallery image $I^{g}_{j}$, we firstly employ the Eq.~\eqref{Eq:S} to compute the similarity with images $I^{q}_{i}$, and denote the similarity as $S_{i,j}$, where $S_{0,j}$ denotes the similarity of probe image $I^{q}_{0}$ and the gallery image $I^{g}_{j}$. 
We then search the nearest neighbor $I^{q}_{N_j}$, which has the maximize similarity score among all neighbors $\boldsymbol{I}^{q}$, where $N_j$ is the index for the nearest neighbor. 
The similar score of the nearest neighbor is denoted as $S_{N_{j},j}$, and $N_j$ can be obtained by:
\begin{equation}
\label{eq:nj}
N_{j}=\arg\max_{i}S_{i,j}.
\end{equation}
Finally, the repulsion term is obtained by taking the gap between similarity of the probe and the nearest neighbor into consideration, defined as the following:
\begin{equation}
S_{gap}(0,j) = S_{0,j}-S_{N_{j},j},
\end{equation}
where $S_{gap}(0,j)$ denotes the difference between the similarity score $S_{0,j}$ and the maximum score $S_{N_{j},j}$. As $I^{q}_{N_j}$ is the nearest neighbor for gallery image $I^{g}_{j}$ in $\boldsymbol{I}^{q}$, $S_{0,j}$ is lower than $S_{N_{j},j}$ that makes $S_{gap}(0,j)\le 0$.  The $S_{gap}(0,j)$ is the absolute difference, which cannot represent the relative difference between two similarities. Therefore,  the repulsion term is obtained by normalizing$S_{gap}(0,j)$ with $S_{N_{j},j}$:
\begin{equation}
\label{Eq:repulsion}
S^{r}(0,j)=e^{\frac{S_{gap}(0,j)}{S_{N_{j},j}}}.
\end{equation}

\begin{figure*}
\begin{center}
\subfigure[]{
\includegraphics[width=0.9\linewidth]{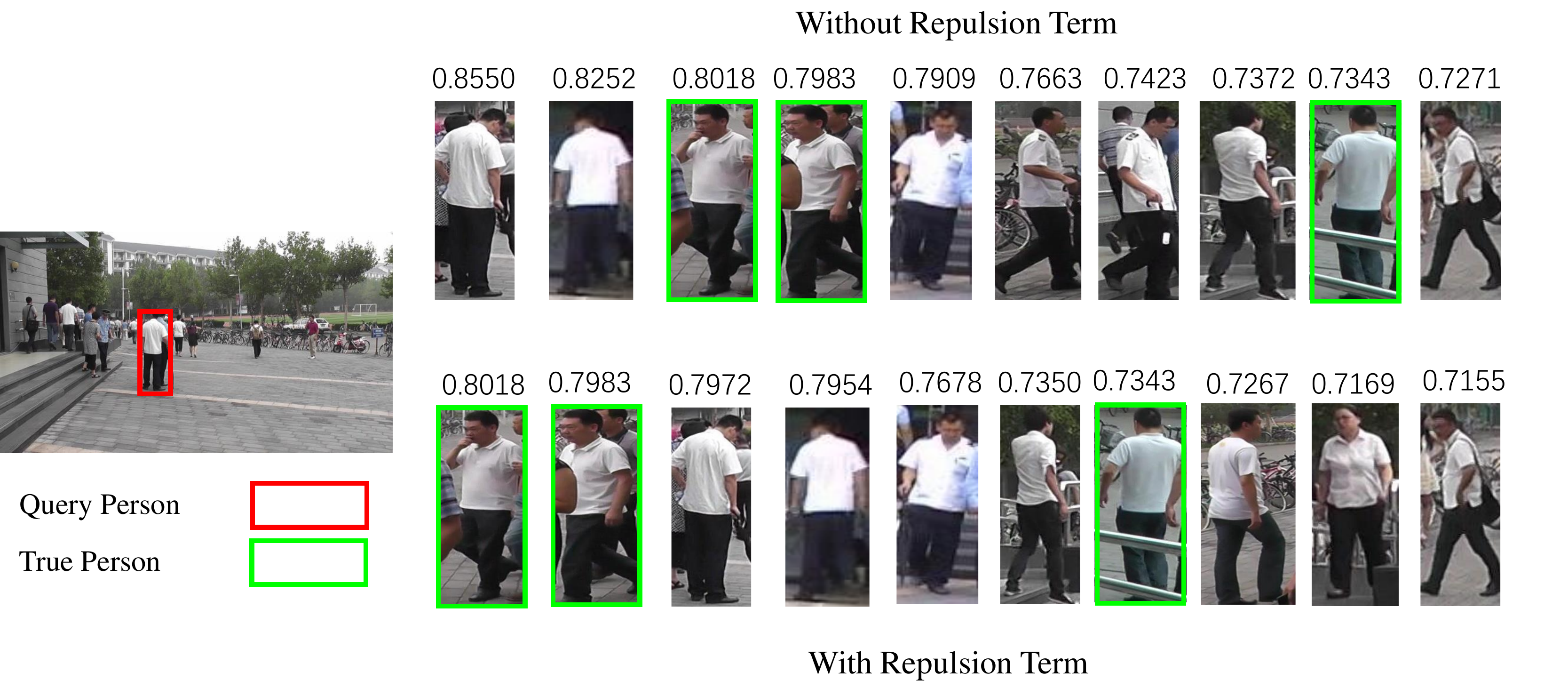}}
\subfigure[]{
\includegraphics[width=0.9\linewidth]{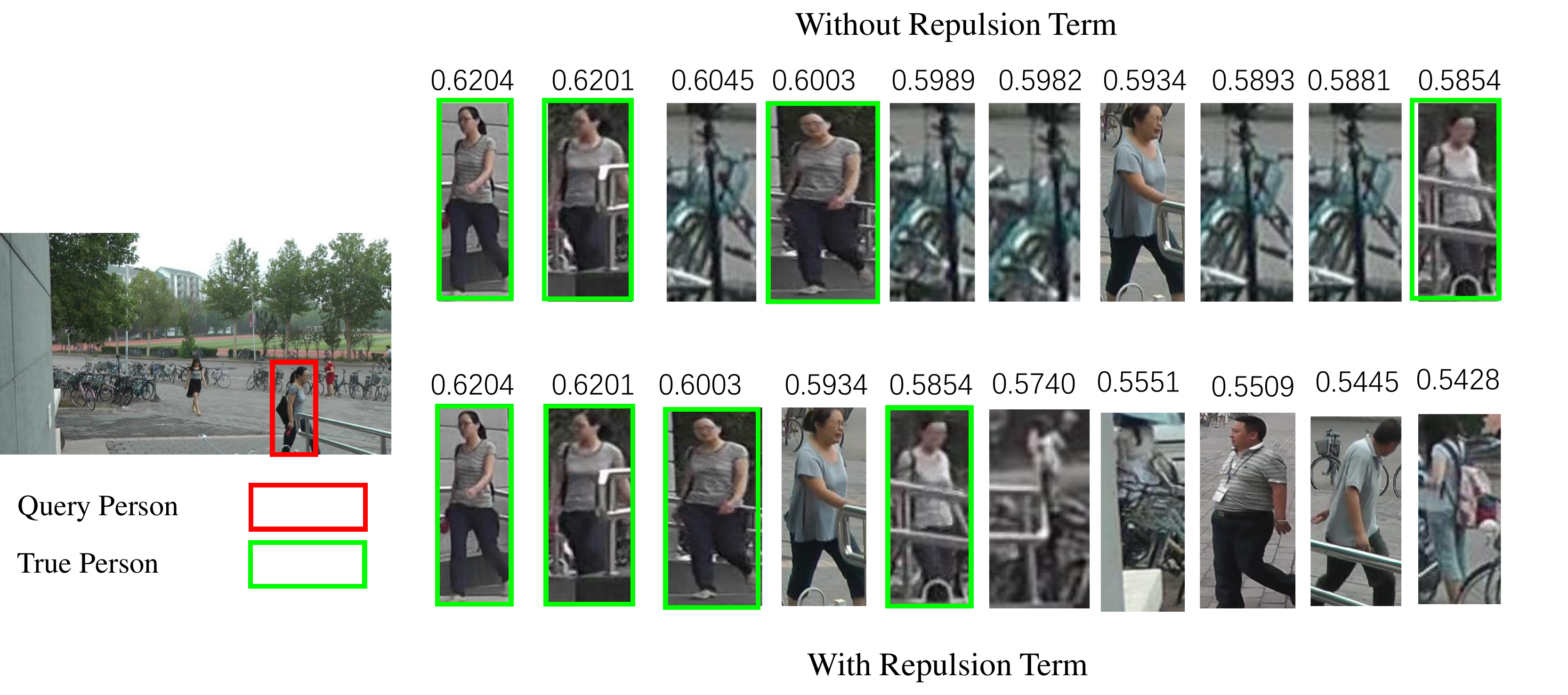}}
\subfigure[]{
\includegraphics[width=0.9\linewidth]{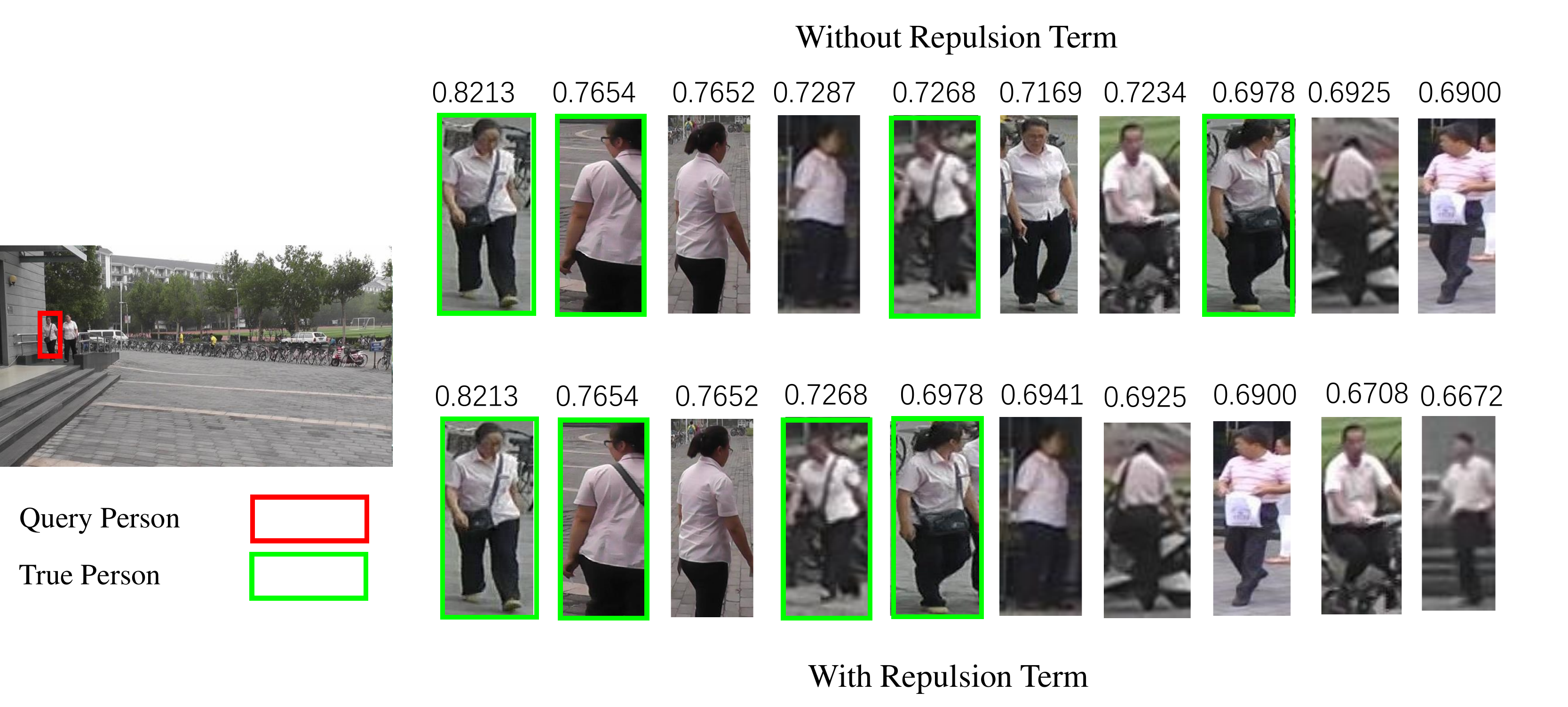}}
\end{center}
\caption{Illustration of the effect of repulsion term. For each sample, the first and second rows illustrate the Top-10 retrieval results without and using the repulsion term, respectively. }
\label{Fig:vis_re}
\end{figure*}

As $S^{r}$ targets to reduce the similarity for negative images, and preserve the similarity for positive images, $S^{r}(0,j)$ has the following characteristics.
Firstly, it does not change the similarity for true positive matching, \emph{e.g.,} for the true positive matching, the nearest neighbor is the probe image itself, thus $N_{j}$=0. 
As a consequence, $S^{r}(0,j)=e^{0}=1$, hence it can preserve the similarity for positive matching. 
Secondly, it needs to reduce the similarity for the gallery image that is most similar to the neighbor $I^{q}_{i}(i\in[1,n_{q}]$). Once the gallery image has a higher similarity with the neighbors,  $S_{gap}(0,j)<0$ due to $ S_{0,j}<S_{N_{j},j}$. The larger gap between  $ S_{0,j}$ and $S_{N_{j},j}$, the lower $S^{r}(0,j)$. Based on the Eq.\eqref{Eq:or},  the lower $S^{r}(0,j)$ can greatly reduce the similarity. Some examples are shown in Figure~\ref{Fig:repul} and Figure~\ref{Fig:vis_re}.

\paragraph{Discussion with CWS} 
The core of Objectness Term and CWS are both to suppress the effect of negative samples during computing the similarity between probe image and gallery images. 
The CWS firstly multiply the normalized confidence score and then calculate the cosine distance between two descriptors. 
Therefore, the weighted similarity is linearly related to the detection score.
Based on the distribution statistics of the detection score shown in Fig.~\ref{Fig:objectness_analysis}, we consider that using linear operation is not a good choice. 
Different from the CWS, the Objectness term proposes an exponent-based weight to fuse the instance, as shown in Eq.~\eqref{eq:ot}.

\paragraph{Discussion with re-ranking method}
The repulsion term is similar to the re-ranking methods, but it has a different motivation with re-ranking methods. 
Since the probe image has a different person ID with its neighbors, the gallery image that has a higher similarity with the neighbors should have a lower similarity with the probe images. 
Therefore, the repulsion term aims to reduce the similarity of the gallery images that is similar to the neighbors of probe person, e.g., as shown in Figure~\ref{Fig:vis_re}, the repulsion term improves the performance of person search by reducing the similar score for distractor images. 
Different from the repulsion term, the re-ranking methods resort to the retrieval results according to the best results that appeared early in the previous retrieval results. 
To obtain the previous retrieval results, the re-ranking methods need to ``see'' all gallery instances. 
Therefore, the re-ranking method is an offline method, which is time-consuming and not suitable for an online person search. 
Different from the offline re-ranking, the repulsion term can be treated as an online method.

\begin{table*}
\begin{center}
\caption{Comparison of results on CUHK-SYSU and PRW with gallery size of 100 and 6,112, respectively.}
\label{Tab:sysu}
\begin{tabular}{l|cc|cc}
\hline
Datasets& \multicolumn{2}{c|}{CUHK-SYSU}& \multicolumn{2}{c}{PRW}\\
\hline
Method & mAP(\%) & Top-1(\%)& mAP(\%) & Top-1(\%) \\
  \hline
\hline
CNN~\cite{RenHG017}+DSIFT~\cite{ZhaoOW13}+Euclidean & 34.5 & 39.4 & & \\
CNN~\cite{RenHG017}+DSIFT~\cite{ZhaoOW13}+KISSME~\cite{KostingerHWRB12} & 47.8 & 53.6 & & \\
CNN~\cite{RenHG017}+IDNet~\cite{XiaoLWLW17} & 68.6 & 74.8 & &\\
  \hline
NPSM~\cite{LiuFJKZQJY17} & 77.9 & 81.2 &24.2&53.1\\
OIM~\cite{XiaoLWLW17} & 75.5 & 78.7 &21.3&49.9\\
SIPR~\cite{LiYC18} & 85.3 & 86.0 &39.5&59.2\\
MGTX~\cite{ChenZOYT18} & 83.0 & 83.7 & 32.6& 72.1\\
CLSA~\cite{LanZG18} & 87.2 & 88.5 &38.7&68.0\\
LCG~\cite{DBLP:conf/cvpr/YanZNZXY19} & 84.1 & 86.5 & 33.4 & 73.6 \\
QEEPS~\cite{DBLP:conf/cvpr/MunjalATG19} & 88.9 & 89.1 & 39.1 & 80 \\
DLR~\cite{han2019re} & 93.0&\textbf{94.2}&42.9&70.2\\
  \hline
Faster R-CNN~\cite{RenHG017}+ResNet50~\cite{HeZRS15} &88.60&89.10&34.17&56.93\\
Faster R-CNN+DenseNet121~\cite{HuangLMW17} &90.13&90.72 &40.42&63.10\\
Faster R-CNN+SeResNet50~\cite{HuSS18} & 88.61&89.45 &42.91&65.05\\
Faster R-CNN+MLFN~\cite{ChangHX18} &88.41&88.62&39.51&62.18\\
Faster R-CNN+PCB~\cite{SunZYTW18} &91.91&92.59&41.95&64.37\\
Faster R-CNN+MGN~\cite{WangYCLZ18} &92.32&93.21&50.91&69.52\\
\hline
Faster R-CNN+ResNet50+\textbf{OR} &90.00&90.21&36.90&59.50\\
Faster R-CNN+DenseNet121+\textbf{OR} & 91.24& 91.93 &41.94&65.58\\
Faster R-CNN+SeResNet50+\textbf{OR} & 89.69&90.24&44.76&68.21\\
Faster R-CNN+MLFN+\textbf{OR} &88.75&89.69&41.51&64.32\\
Faster R-CNN+PCB+\textbf{OR} &92.93&93.69&43.01&65.87\\
Faster R-CNN+MGN+\textbf{OR} &\textbf{93.23}&93.83&\textbf{52.30}&\textbf{71.51}\\
\hline
\end{tabular}
\end{center}
\end{table*}

\section{Experiments}
To demonstrate the effectiveness of the OR Similarity, several experiments are conducted on the two widely used datasets~\footnote{Currently, there are only two datasets, CUHK-SYSU and PRW, for person search task.}.
\subsection{Datasets}
\paragraph{CUHK-SYSU} CUHK-SYSU~\cite{XiaoLWLW17} is a large-scale person search dataset consisting of street/urban scene images captured by camera or movie snapshots. It contains 18,184 scene images with 96,143 annotated bounding boxes. Among all the bounding boxes, there are 8,432 labeled person IDs, and each person appears in two or more different gallery images. The all bounding boxes and person IDs are split into two groups: 1) the training set contains 11,206 images and 5,532 person IDs; 2) the testing set has 6,978 images with 2,900 person IDs. There are several subsets with different gallery sizes for each probe person. During the evaluation, each subset is officially defined by the dataset with a size of 50, 100, 500, 1000, 2000, 4000.

\paragraph{PRW} Different from the CUHK which are camera-independently, the Person Re-identification in the Wild(PRW)~\cite{ZhengZSCYT17} dataset is collected from the video frames that are captured by six cameras. Therefore, PRW is more challenging than the CUHK-SYSU dataset. There are 43,110 annotated bounding boxes generated from 11,816 video frames. Based on the standard training/test split provided by the dataset, the training dataset contains 482 person IDs from 5,704 video frames, and the test set has 2,057 person IDs with 6,112 gallery images. The standard evaluation of PRW takes the whole gallery set into consideration while searching each probe person. Besides using the full gallery set, we also randomly generate some subsets with gallery size of 200, 500, 1000, 2000, and 4000, used for evaluation.

\subsection{Evaluation Metrics}
For the person detection, a detection box is considered as a correct matching if the intersection-over-union (IoU) with any grounding box is above 0.5. Similar to object detection, Average Precision (AP) and recall are used to measure the performance of person detection. For the person search, the cumulative matching characteristics (CMC top-K) metric and the mean averaged precision (mAP) are employed to measure the performance of the person search system. The CMC top-K computes a matching as there is at least one of the top-K predicted bounding boxes with the intersection-over-union (IoU) larger than or equal to 0.5. The mAP is used to reveal the accuracy and recall rate for each probe image.

\subsection{Implementation Details}
In this work, all models and evaluations are performed with the Pytorch framework. The person detector is implemented based on the Faster R-CNN with the backbone of ResNet-101, and initialized with a COCO-pretrained model~\footnote{https://github.com/jwyang/faster-rcnn.pytorch}. 
The Faster R-CNN is trained with two-step training strategies with the detection loss introduced original work. 
We firstly train the Faster R-CNN on the labeled person bounding boxes among COCO datasets, and then fine-tune the trained Faster R-CNN on person search datasets, \emph{e.g.,} CUHK-SYSU, and PRW. 
The detector training is performed with the SGD algorithm with the momentum setting to 0.9, the weight decay to 0.0001, the iteration to three epochs with the learning rate of 0.001.

For the person description, we employ six type of models to generate the representation, \emph{i.e.,} ResNet50\cite{HeZRS15}, DenseNet121\cite{HuangLMW17}, Squeeze-and-Excitation Networks (SE)\cite{HuSS18}, Multi-Level Factorisation Net (MLFN)\cite{ChangHX18}, Part-based Convolutional Baseline (PCB)\cite{SunZYTW18}, Multiple Granularities Networks (MGN)\cite{WangYCLZ18}~\footnote{The MGN is one of the state-of-the-art single models for person re-identification.}. 
The ResNet50, DenseNet121, and Se-Resnet50 are the traditional image classification models, and the MLFN, PCB, and MGN are the person re-identification models. 
Except for the MGN, the rest models are all trained based on the public implementation~\footnote{https://github.com/KaiyangZhou/deep-person-reid} with the following setting: 1) using the classification loss and triplet loss. 2) the learning rate is set as 0.003 with the stepsize of 50. For the MGN, we employ the pytorch implementation~\footnote{https://github.com/seathiefwang/MGN-pytorch} with its default setting. Targeting to train a robust person description model, we employ the labeled person images as the training images, and all person bounding boxes are resized to $384\times 128$.

\begin{figure*}
\begin{center}
\subfigure[]{
\includegraphics[width=0.32\linewidth]{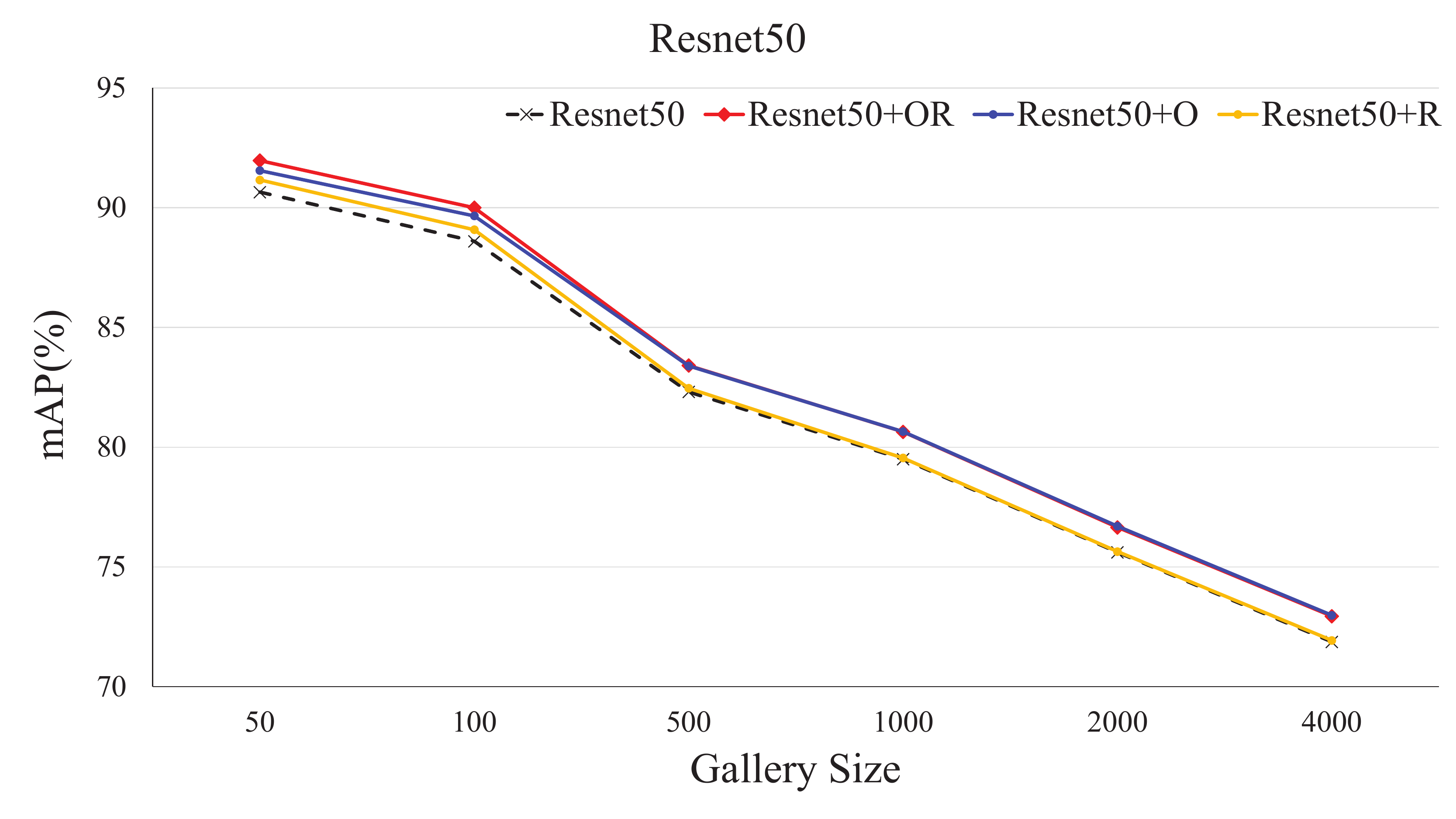}}
\subfigure[]{
\includegraphics[width=0.32\linewidth]{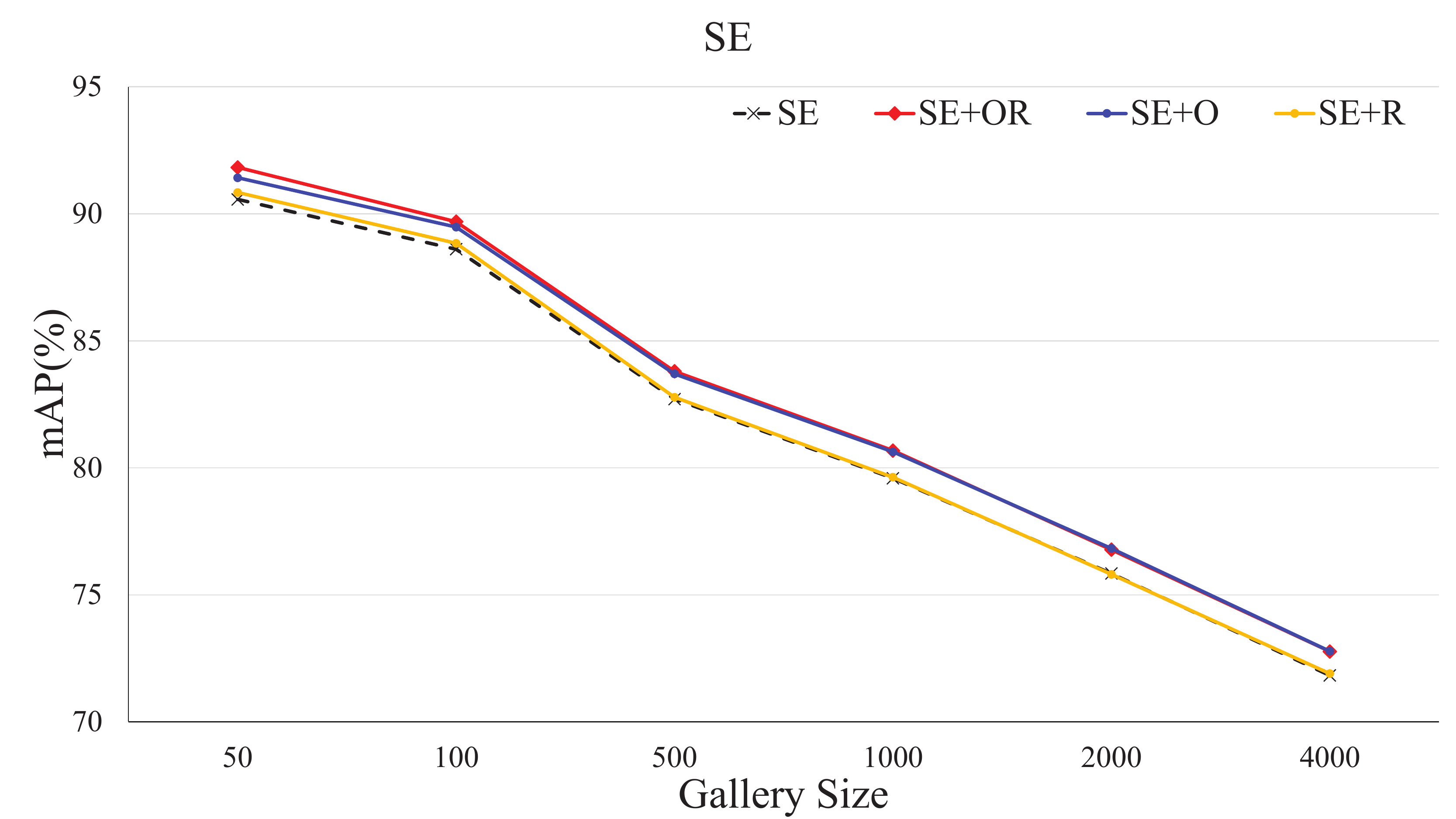}}
\subfigure[]{
\includegraphics[width=0.32\linewidth]{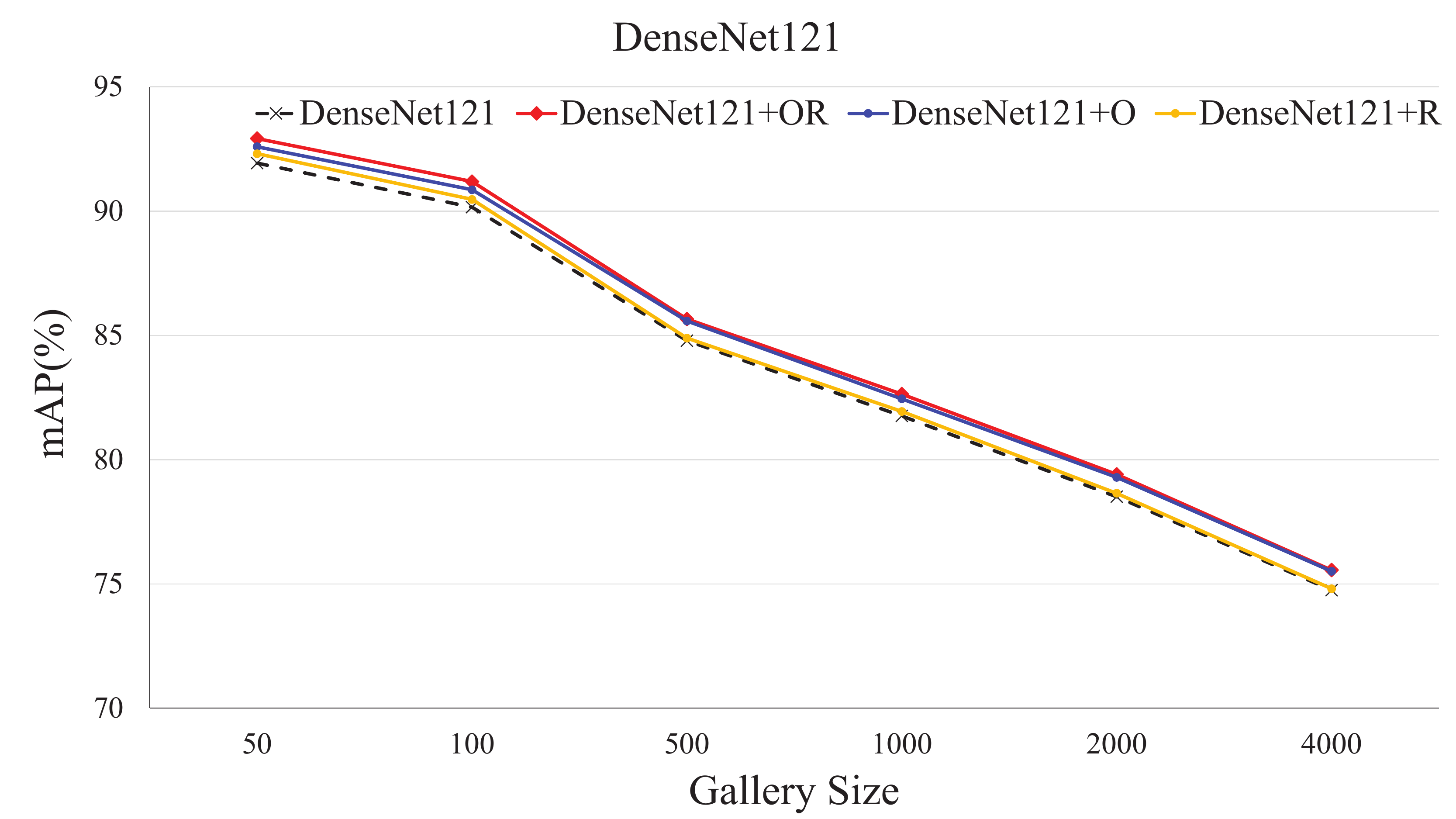}}\\
\subfigure[]{
\includegraphics[width=0.32\linewidth]{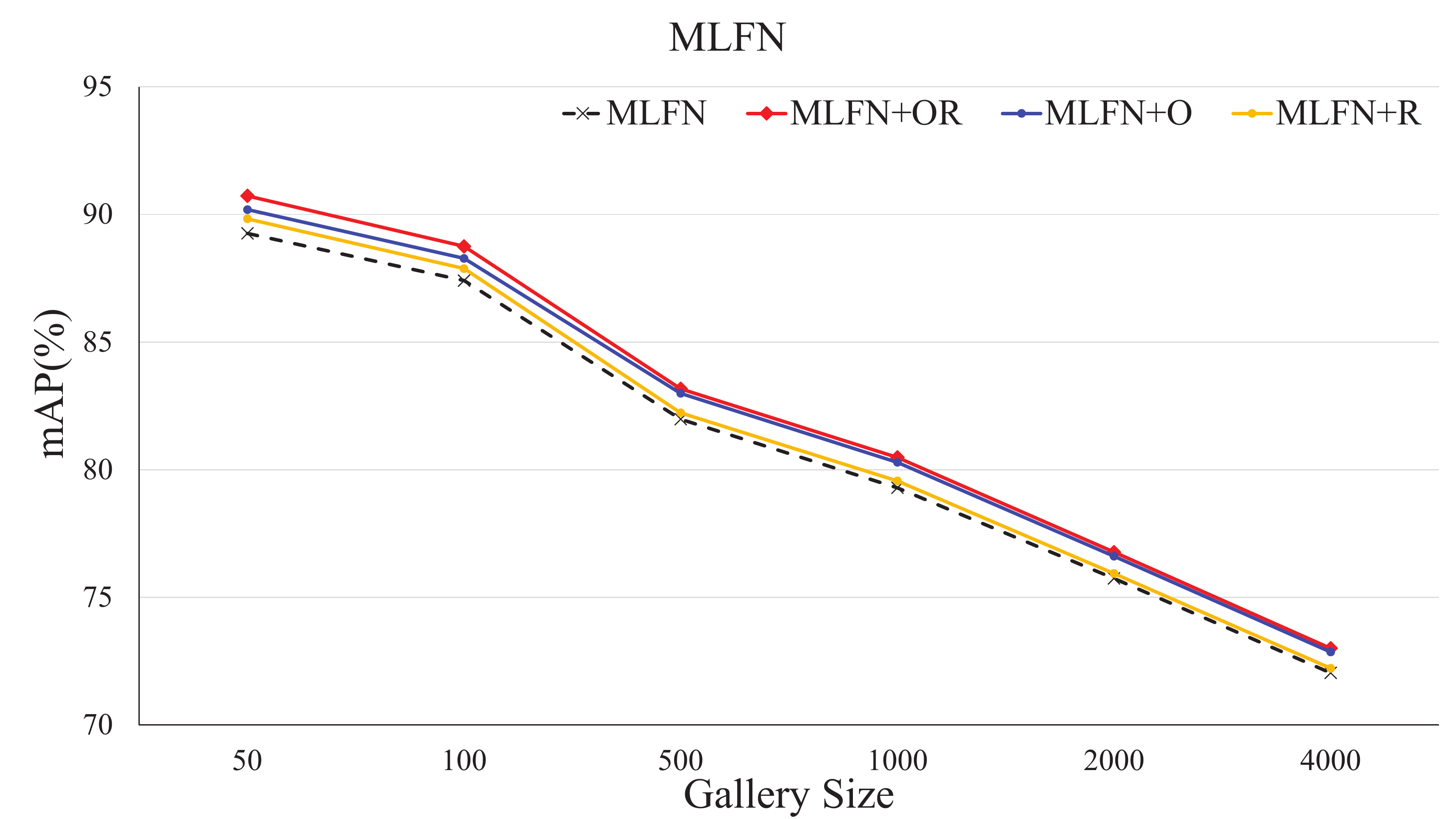}}
\subfigure[]{
\includegraphics[width=0.32\linewidth]{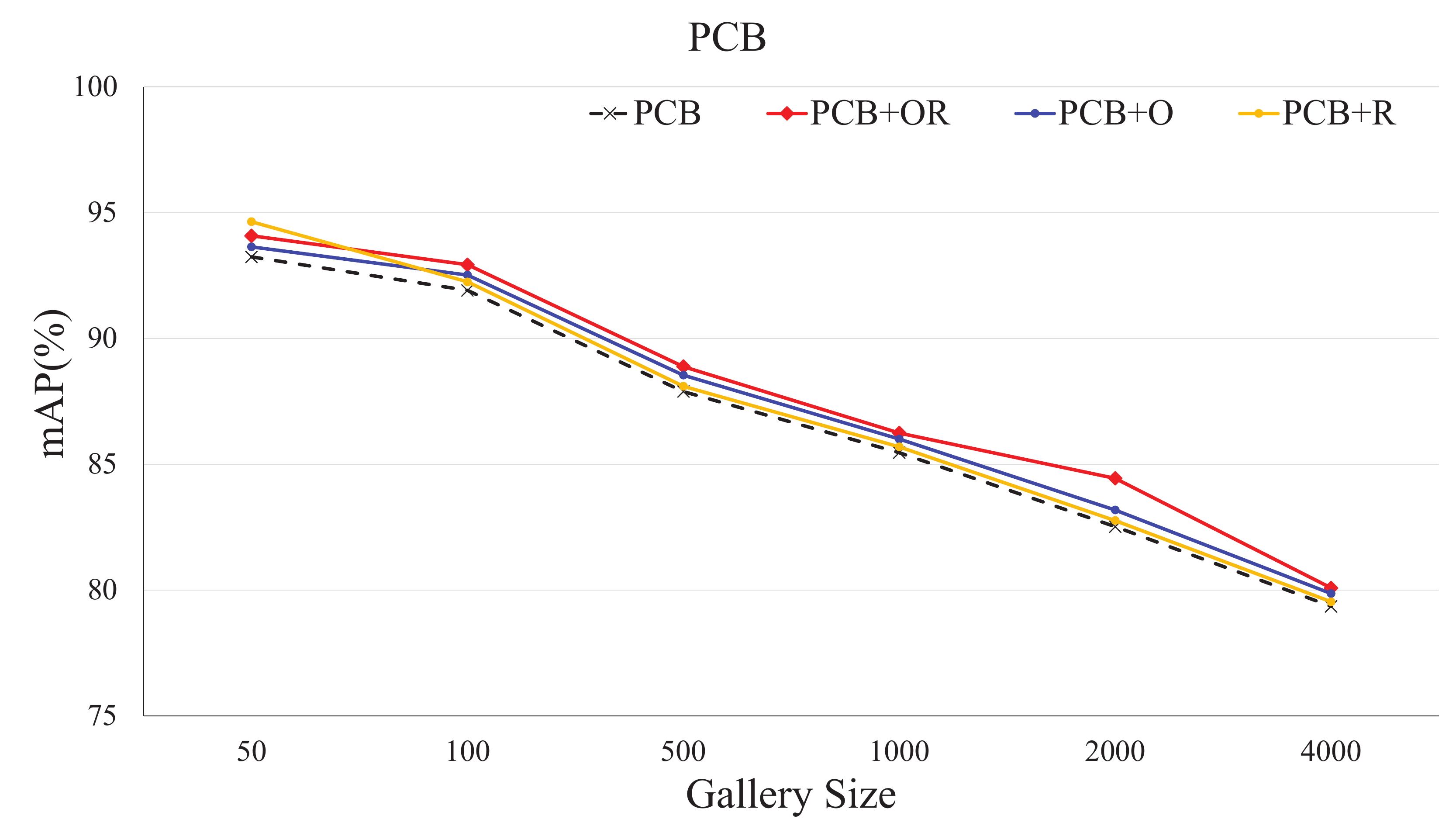}}
\subfigure[]{
\includegraphics[width=0.32\linewidth]{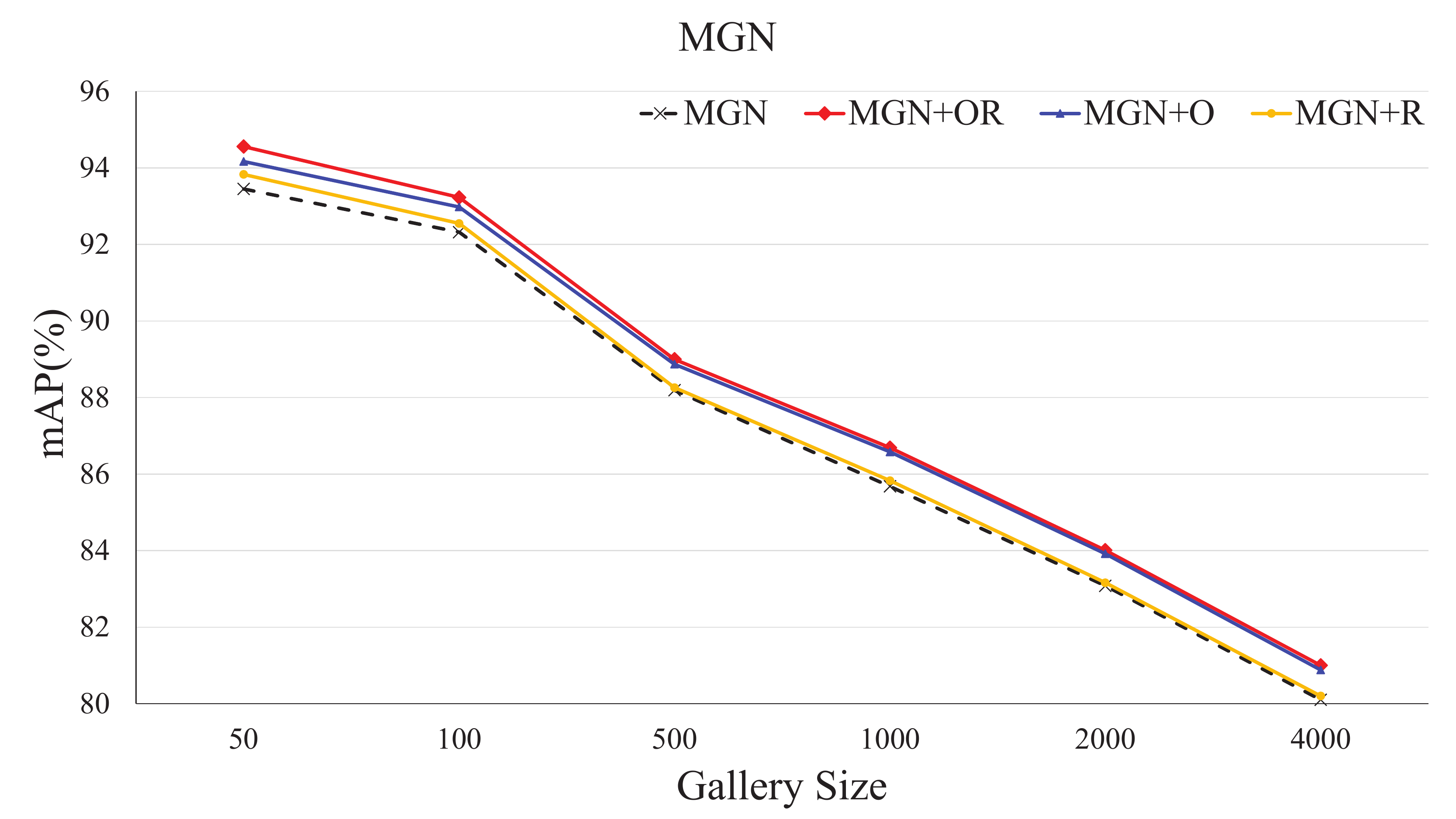}}
\end{center}
\caption{The effect of objectness term (O), repulsion term (R), and OR similarity (OR) on the CUHK-SYSU dataset with varying gallery sizes.}
\label{Fig:CUHKOr}
\end{figure*}

\subsection{Comparision with State-of-The-Art Methods}
\begin{figure}
\includegraphics[width=1\linewidth]{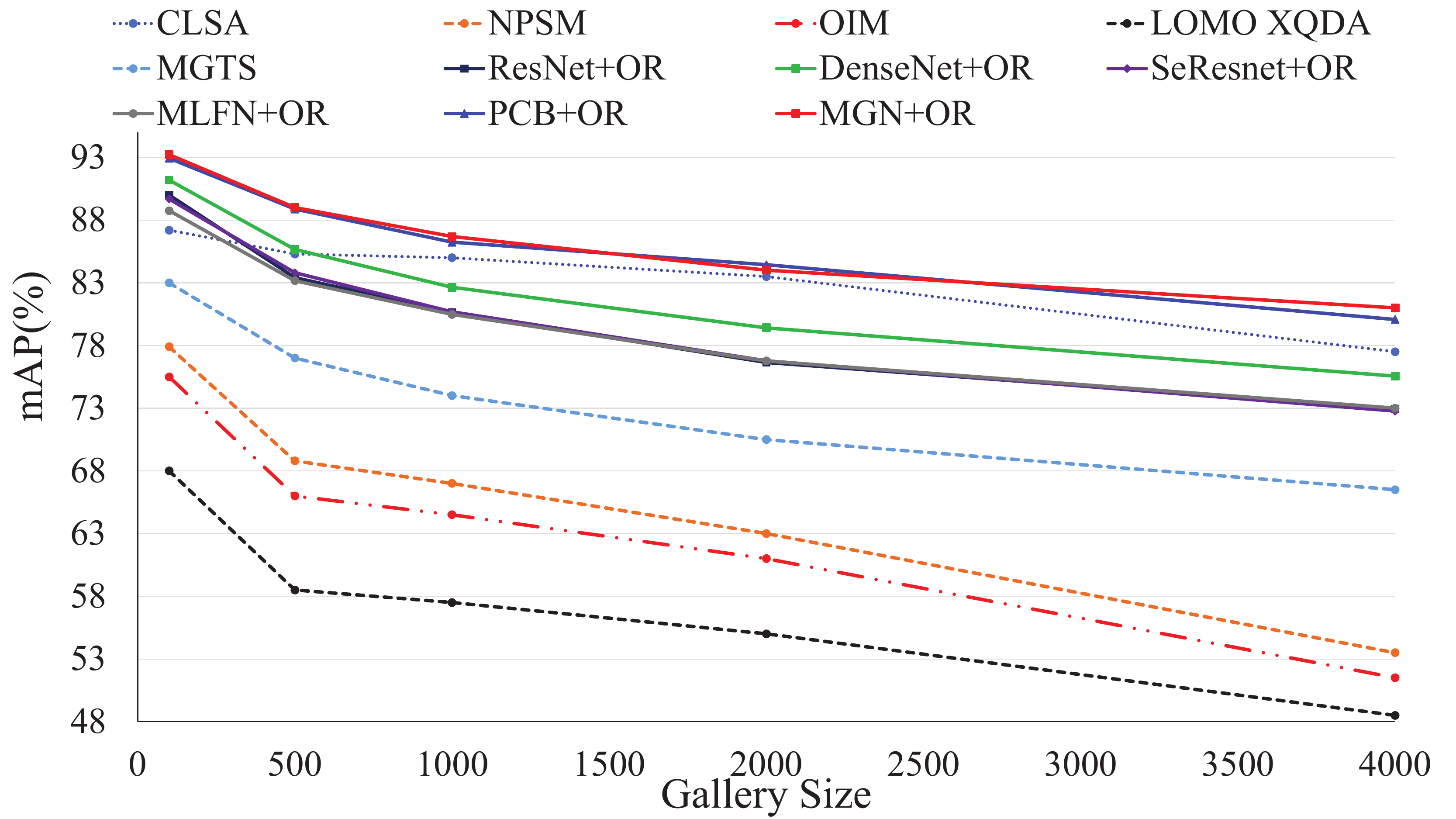}
\caption{Performance comparison on CUHK-SYSU with varying gallery sizes.}
\label{Fig:sysu_cmp}
\end{figure}

In this section, we report the person search results on the CUHK-SYSU and PRW datasets, with a comparison with several state-of-the-art methods, \emph{e.g,} OIM~\cite{XiaoLWLW17}, NPSM~\cite{LiuFJKZQJY17}, MGTS~\cite{ChenZOYT18}, CLSA~\cite{LanZG18}, LCG~\cite{DBLP:conf/cvpr/YanZNZXY19}, QEEPS~\cite{DBLP:conf/cvpr/MunjalATG19}, and DLR~\cite{han2019re}. The related results are summarized in Figure~\ref{Fig:sysu_cmp} and Table~\ref{Tab:sysu}.

Firstly, we show that we have been implemented a robust baseline framework for the person search. 
As shown in Table~\ref{Tab:sysu}, without using OR similarity, most of the models by using the detection results from Faster R-CNN obtain the better performance than existing methods expect of the DLR~\cite{han2019re}. 
The reason is that the DLR proposes a more robust person detector than Fast R-CNN used in ours, \emph{e.g.,} for the CUHK-SYSU dataset, the detector in DLR obtains the mAP of 93.09\%, which is higher than the mAP of 87.78\% of ours.
Therefore, the more robust detector makes the DLR obtain a higher performance than ours.
From the Table~\ref{Tab:sysu}, we also observe that the QEEPS~\cite{DBLP:conf/cvpr/MunjalATG19} and CLSA~\cite{LanZG18} both obtain a higher performance than the \emph{Faster R-CNN+ResNet50}.  
The reason is that those two methods employ the more robust description model than ResNet50.

We then show that adding the OR term upon the existing methods can further boost their performance. 
As shown in Table~\ref{Tab:sysu}, using OR term obtains the improvement of mAP from 1.08\% to 3.16\% upon existing methods for those two datasets.
Once using the OR similarity, most methods achieve a better performance than existing methods.

Finally, combining the MGN with OR term achieves the current state-of-the-art performance, \emph{e.g.,}  the MGN obtains the mAP of 0.23\% and 9.4\% improvement over the existing highest performance for CUHK-SYSU and PRW datasets, respectively. 
Figure~\ref{Fig:sysu_cmp} further illustrates the comparison of the CUHK-SYSU dataset with different gallery sizes. 
The better performance demonstrates that the proposed OR similarity is compatible with existing methods and can further boost the performance.

\subsection{Ablation Study}
From the above experimental results, we can observe that using the OR term can achieve significant improvement over the six models on both two datasets. 
Therefore, OR similarity is a useful measurement for person search. 
The significant contribution of the OR term is that it takes the objectness of gallery images, and repulsion between the probe image and its neighbors into consideration. 
To understand the impact of the above two aspects, we conduct some studies on those two datasets with different gallery sizes.

\begin{figure*}
\begin{center}
\subfigure[]{
\includegraphics[width=0.31\linewidth]{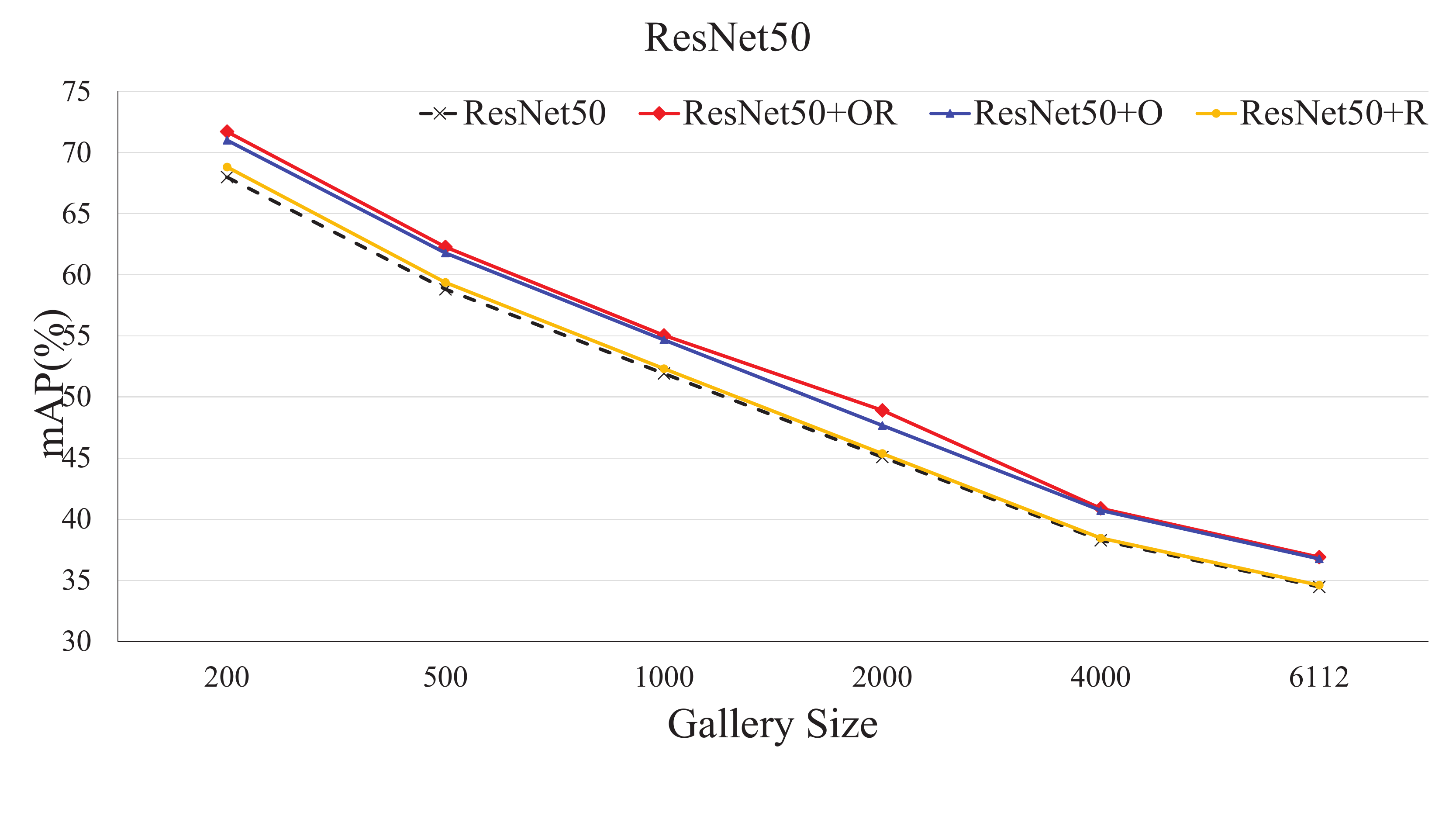}}
\subfigure[]{
\includegraphics[width=0.31\linewidth]{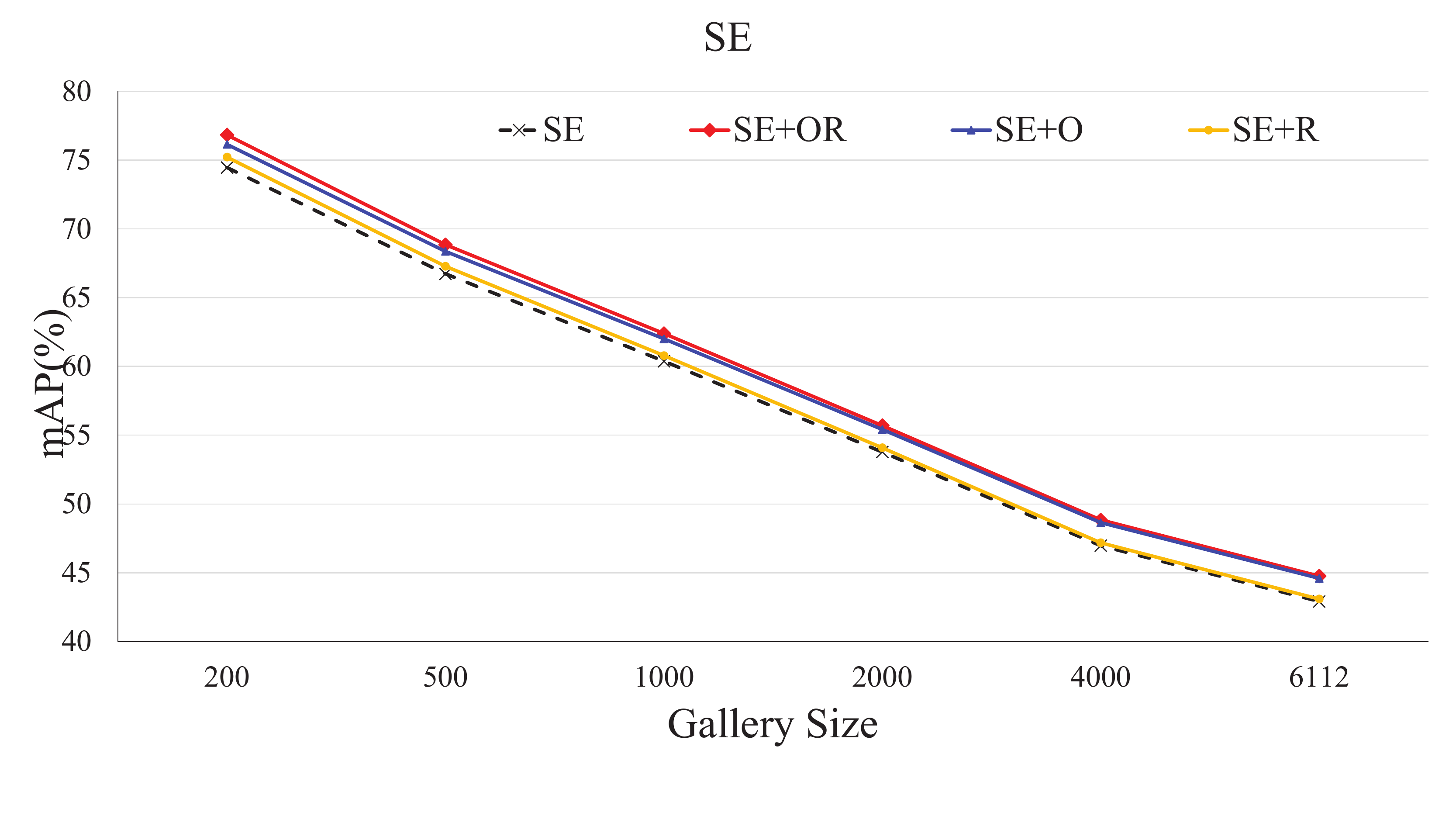}}
\subfigure[]{
\includegraphics[width=0.31\linewidth]{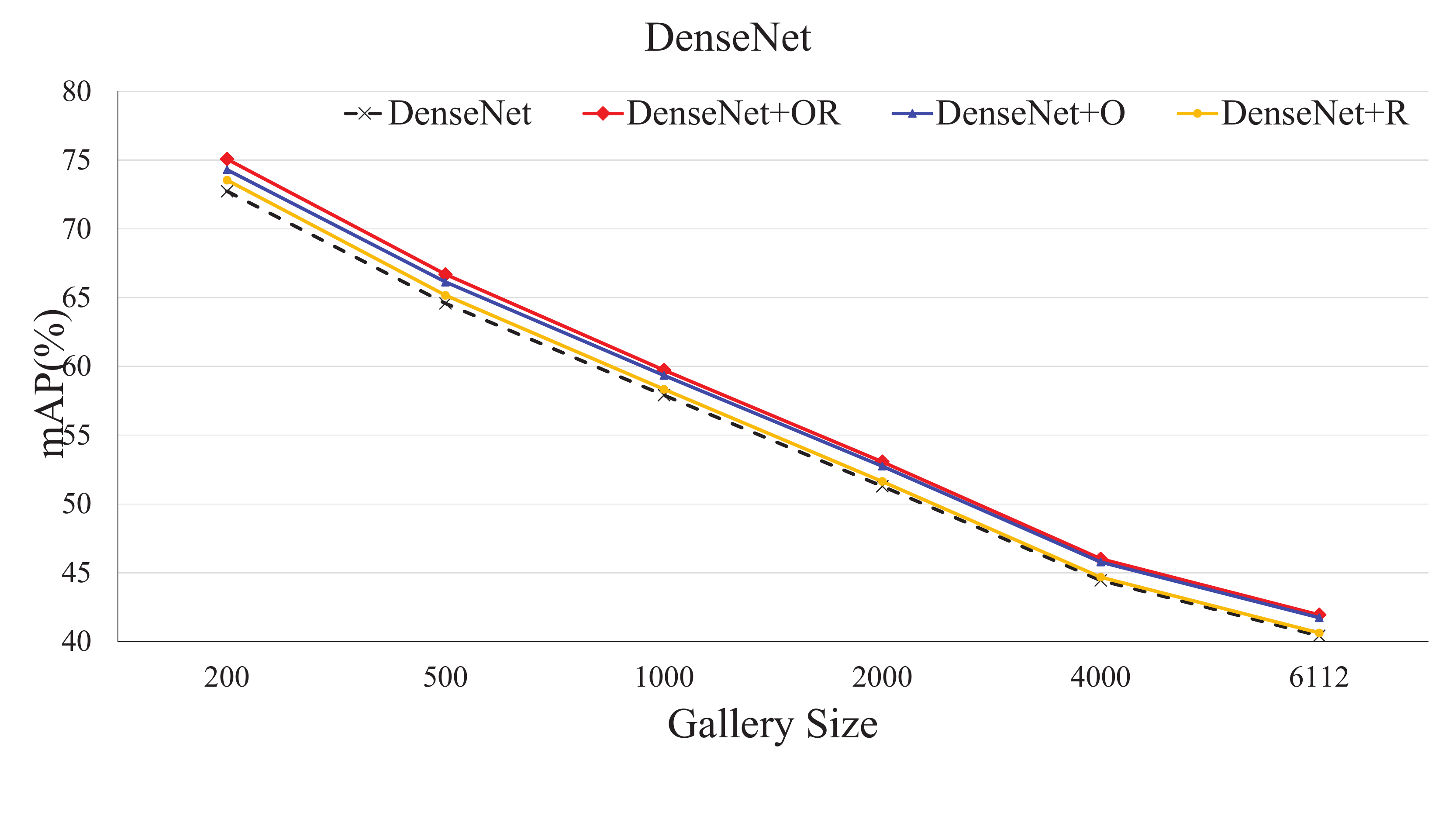}}\\
\subfigure[]{
\includegraphics[width=0.31\linewidth]{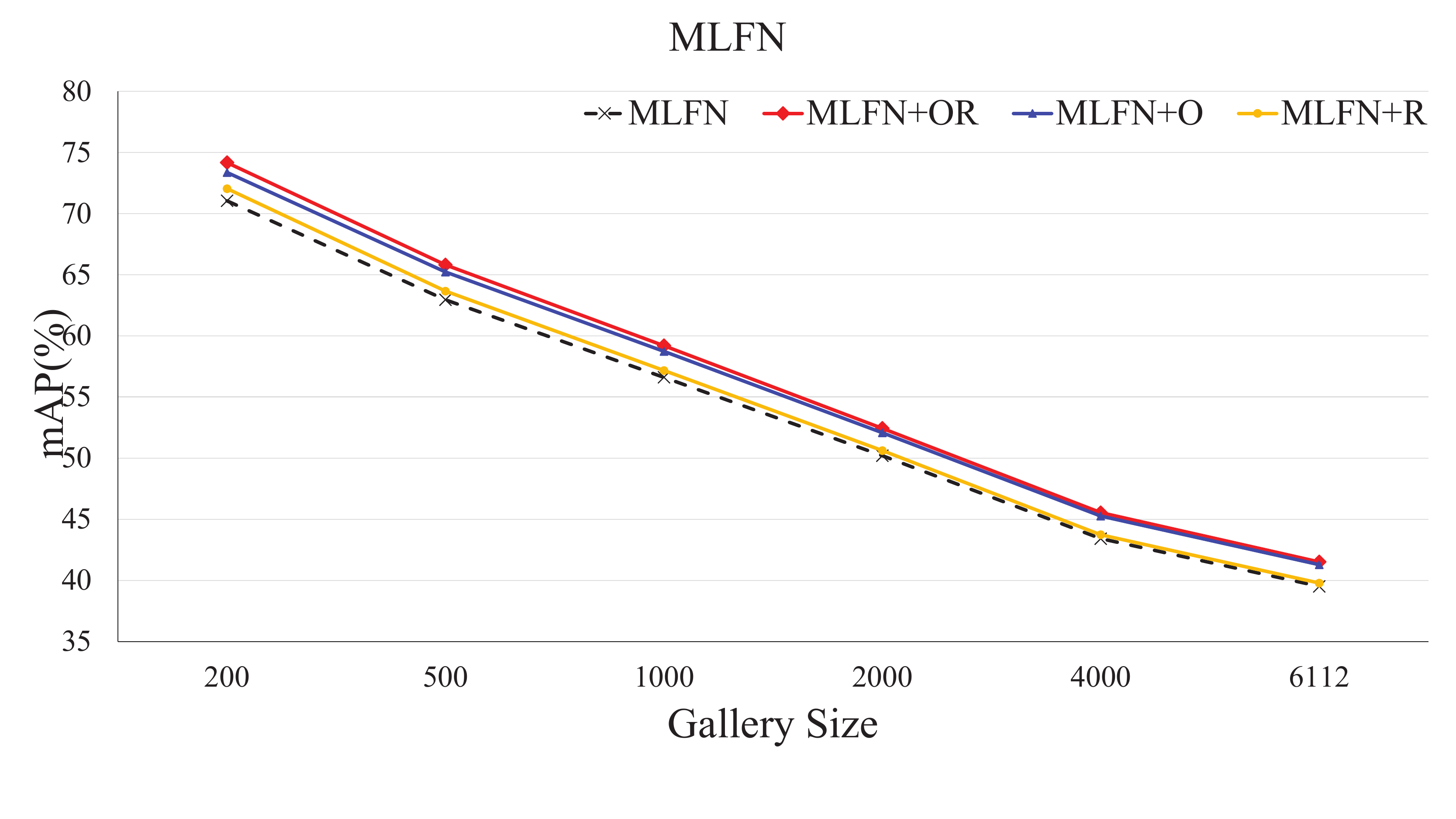}}
\subfigure[]{
\includegraphics[width=0.31\linewidth]{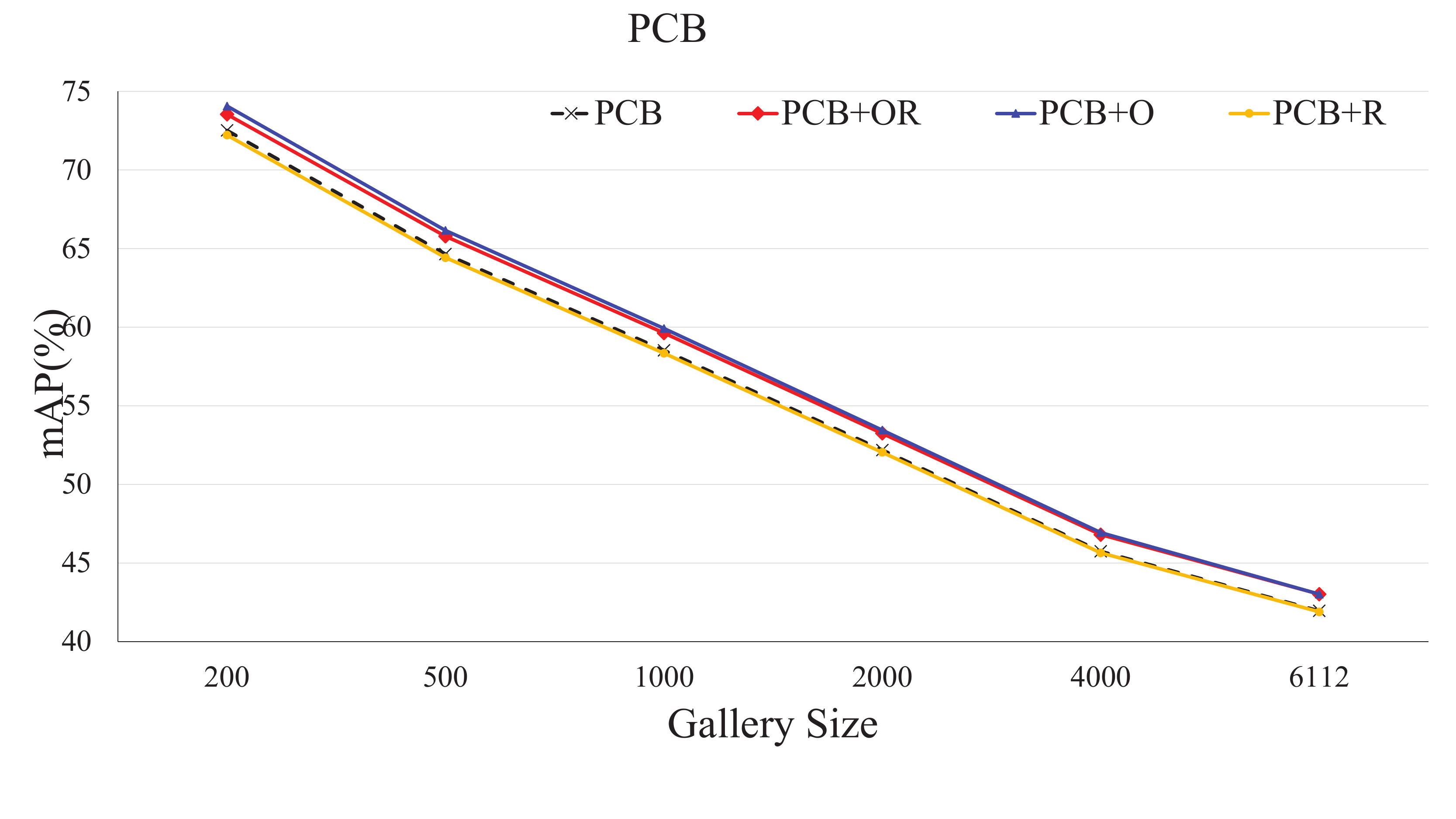}}
\subfigure[]{
\includegraphics[width=0.31\linewidth]{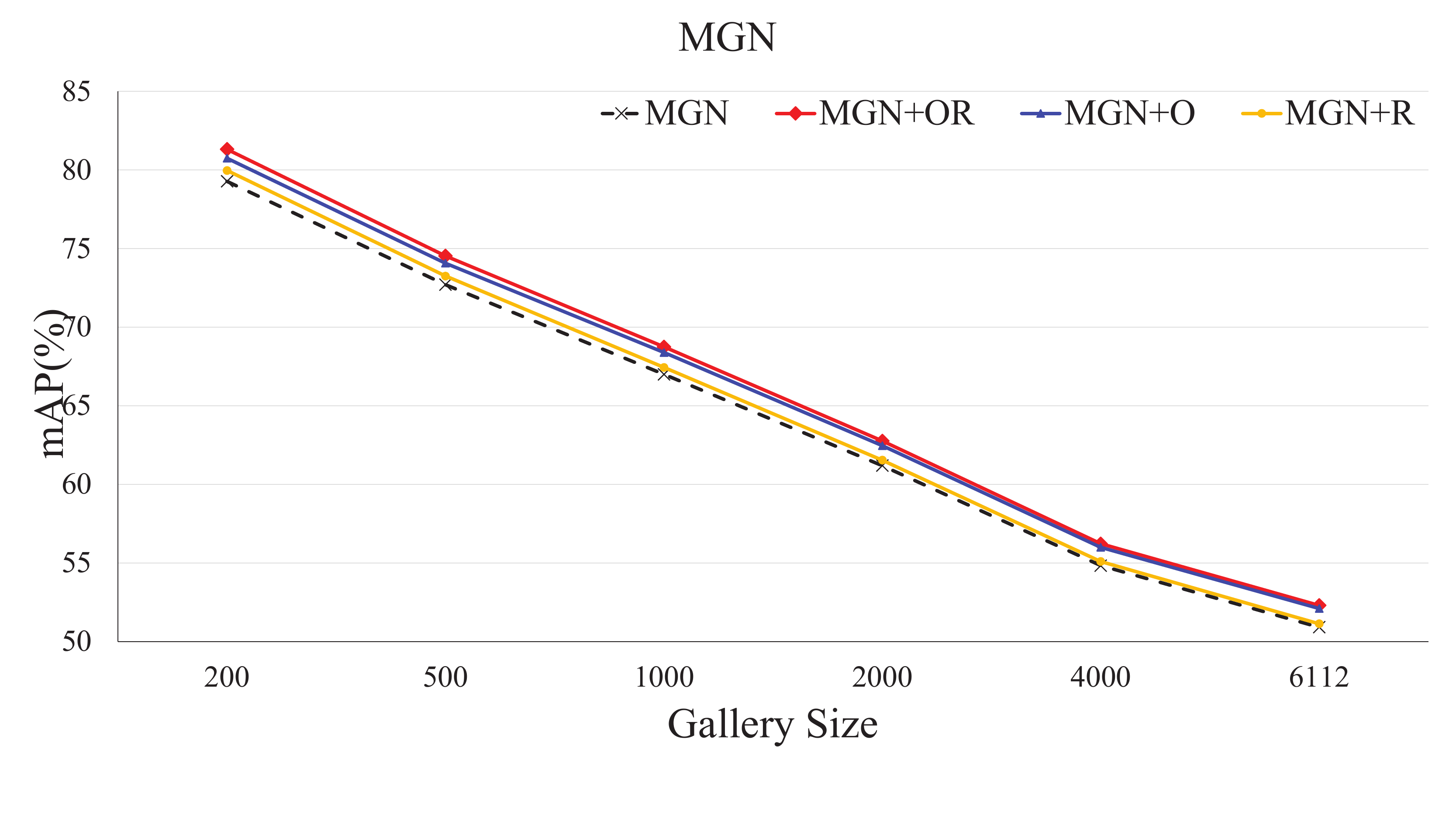}}
\end{center}
\caption{The effect of objectness term (O), repulsion term (R), and OR similarity (OR) on the RPW dataset with varying gallery sizes.}
\label{Fig:PRWOr}
\end{figure*}

\paragraph{Person Detection}
As the person detection plays a vital role in the person search, we then report the results for the detector implemented in this paper. The detailed results are shown in Table~\ref{Tab:det}. As shown in Table~\ref{Tab:det}, we can see that the detector used in this work obtains higher performance than the existing CLSA. The robust detector can generate much accurate person proposals used for person matching.

\begin{table}
\caption{The detection performance for CUHK-SYSU and PRW dataset.}
\label{Tab:det}
\begin{tabular}{lccc}
\hline
Datasets & Methods & mAP(\%) & Recall(\%)\\
\hline
\hline
CUHK-SYSU & CLSA~\cite{LanZG18} & 87.2 & 88.5 \\
CUHK-SYSU & Ours & 87.78 & 91.81 \\
PRW &Ours& 91.36 & 96.33 \\
\hline
\end{tabular}
\end{table}

\begin{figure}
\begin{center}
\includegraphics[width=0.9\linewidth]{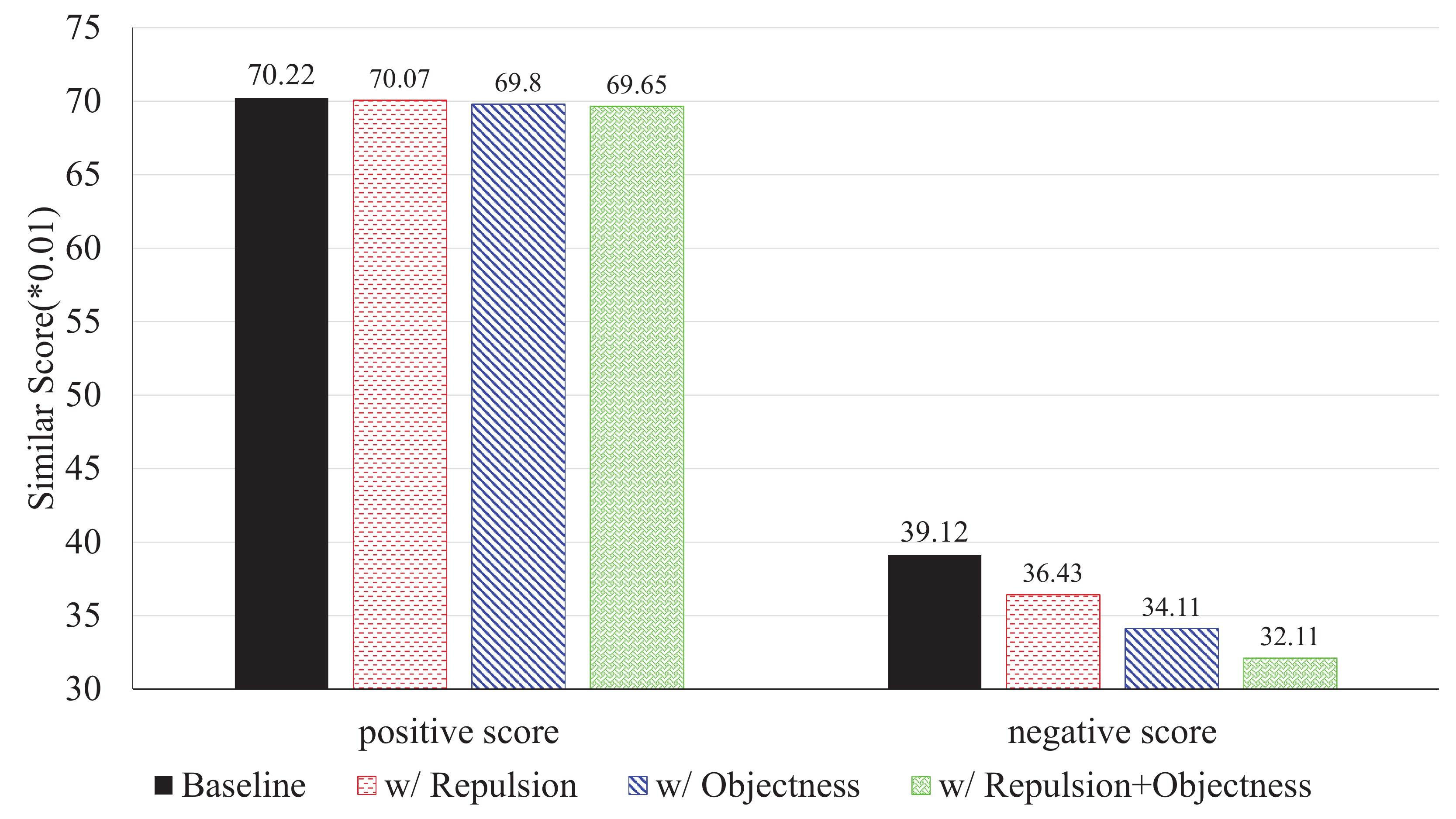}
\end{center}
\caption{The effect of the OR similarity on the similarity of positive and negative images on the PRW dataset with the gallery size of 200.}
\label{Fig:effect}
\end{figure}

\paragraph{Effect of Objectness term} We evaluate the effect of objectness term for person search, and summarize the related results for CUHK-SYSU and PRW in Figure~\ref{Fig:CUHKOr} and Figure~\ref{Fig:PRWOr}, respectively. From the Figure~\ref{Fig:CUHKOr} and Figure~\ref{Fig:PRWOr}, it can be observed that using the objectness constraint can achieve improvement for both large scale and small scale gallery sizes. 
Since the objectness term is proposed to reduce similar scores for distractor images, we further illustrate the similar score affected by the objectness term. 
As shown in Figure~\ref{Fig:effect}, using objectness term produces a significant influence on the similarity of the distractor images and maintains the similarity of the positive images. 
For example,  the objectness term can reduce the similarity of positive and negative images from 0.7032 and 0.3912 to 0.698 and 0.3411, respectively. 
Therefore, we can conclude that using the objectness term can reduce the effect of distractor images for person search.

\begin{table*}
\begin{center}
\caption{Effect of objectness term (O), repulsion term (R), and OR term on the CUHK-SYSU dataset with gallery size of 50 on six different description models.}
\label{Tab:effect_om}
\begin{tabular}{lcccccc}
\hline
Methods & ResNet50 & SE-Net & DenseNet121 & MLFN & PCB & MGN\\
\hline
\hline
MGN& 90.65 & 90.57 & 91.93 & 89.26& 91.91 & 93.45\\
MGN+R &91.16 &90.84& 92.31 & 89.83 & 92.24 & 93.83\\
MGN+O & 91.55& 91.42 & 92.59 & 90.19 & 92.52 & 94.17\\
MGN+OR & 91.97 & 91.83 & 92.92 & 90.73 & 92.93 & 94.56\\
\hline
\end{tabular}
\end{center}
\end{table*}

\begin{table*}
\begin{center}
\caption{Effect of objectness term (O), repulsion term (R), and OR term on the PRW dataset with gallery size of 200 on six different description models.}
\label{Tab:effect_om_prw}
\begin{tabular}{lcccccc}
\hline
Methods & ResNet50 & SE-Net & DenseNet121 & MLFN & PCB & MGN\\
\hline
\hline
MGN& 67.99 & 74.46 & 72.74 & 71.05& 72.52 & 79.28\\
MGN+R &68.81&75.22& 73.55 & 72.04 & 72.22 & 79.97\\
MGN+O & 71.01& 76.15 & 74.31 & 73.37 & 74.07 & 80.75\\
MGN+OR & 71.71& 76.84 & 75.08 & 74.18 & 73.54 & 81.32\\
\hline
\end{tabular}
\end{center}
\end{table*}

\paragraph{Effect of Repulsion term} We then evaluate the effectiveness of repulsion term for person search. 
From the Figure~\ref{Fig:CUHKOr}, Figure~\ref{Fig:PRWOr}, Table~\ref{Tab:effect_om}, and Table~\ref{Tab:effect_om_prw}, we can see that simply using the repulsion term can improve the performance. 
However, the effect of repulsion term is related to the size of gallery, \emph{e.g.,} for the MGN model, the improvement of repulsion term for the size of 200, 500, 1000, 2000, 4000, 6112 are 0.69\%, 0.55\%, 0.44\%, 0.35\%, 0.26\%, 0.22\% on PRW dataset, respectively. 
The reason is that the repulsion term is used to adjust the distractor image that has a higher similarity with the probe image, and the larger gallery may contain many distractor images that are not similar to the probe images. 
The more gallery images, the smaller the percentage of gallery images can be adjusted by the repulsion term. 
We further demonstrate the impact of repulsion term on the similarity score and give more visualization results. 
As shown in Figure~\ref{Fig:effect} and Figure~\ref{Fig:vis_re}, using repulsion term does not destroy the similarity of positive images, but dramatically reduces the score of negative images. 
Finally, we provide some statistical results to prove that the repulsion term is a useful term for the person search. 
By comparing the performance between without using repulsion term and using repulsion term, we found that using the repulsion term achieves performance improvement on 49.2\% of query samples, and only reduces performance on 11.52\% of samples.
Therefore, the repulsion term has a positive effect on half of the probe images.
Therefore, using a repulsion term can effectively reduce the impact of distractor images on the person search.

\paragraph{Effect of OR term} As the objectness and repulsion are two independent terms, combining the objectness  and repulsion term can further boost the performance. 
From the Figure~\ref{Fig:CUHKOr}, Figure~\ref{Fig:PRWOr}, Table~\ref{Tab:effect_om}, and Table~\ref{Tab:effect_om_prw}, , we can see  that using the OR term obtains a higher performance than merely using objectness or repulsion terms. 
Figure~\ref{Fig:effect} shows that using the OR term can further reduce the score of negative images.

\section{Conclusion}
Different from the existing person search methods, our work focuses on the person similarity measurement for person search. By taking the repulsion of probe image, and the objectness of gallery images into consideration, we propose an effective similarity to consider the person objectness and repulsion (OR similarity). 
The OR similarity improves the performance for person search by reducing the effect of distractor images. The experiments on six description models demonstrate the effectiveness of the proposed method. This work has been shown that using the repulsion between probe person and its neighbors can help to recognize the person in the case of occlusion. Although the repulsion term is non-parametric, it relies heavily on the robust of the person description. In the future, we will investigate how to combine the repulsion constraint into the representation learning, and make the generated representation contain the repulsion information.

\ifCLASSOPTIONcaptionsoff
  \newpage
\fi

\bibliographystyle{IEEEtran}
\bibliography{IEEEabrv,egbib}

\begin{thebibliography}{10}
\providecommand{\url}[1]{#1}
\csname url@samestyle\endcsname
\providecommand{\newblock}{\relax}
\providecommand{\bibinfo}[2]{#2}
\providecommand{\BIBentrySTDinterwordspacing}{\spaceskip=0pt\relax}
\providecommand{\BIBentryALTinterwordstretchfactor}{4}
\providecommand{\BIBentryALTinterwordspacing}{\spaceskip=\fontdimen2\font plus
\BIBentryALTinterwordstretchfactor\fontdimen3\font minus
  \fontdimen4\font\relax}
\providecommand{\BIBforeignlanguage}[2]{{%
\expandafter\ifx\csname l@#1\endcsname\relax
\typeout{** WARNING: IEEEtran.bst: No hyphenation pattern has been}%
\typeout{** loaded for the language `#1'. Using the pattern for}%
\typeout{** the default language instead.}%
\else
\language=\csname l@#1\endcsname
\fi
#2}}
\providecommand{\BIBdecl}{\relax}
\BIBdecl

\bibitem{XiaoLWLW17}
\BIBentryALTinterwordspacing
T.~Xiao, S.~Li, B.~Wang, L.~Lin, and X.~Wang, ``Joint detection and
  identification feature learning for person search,'' in \emph{2017 {IEEE}
  Conference on Computer Vision and Pattern Recognition, {CVPR} 2017, Honolulu,
  HI, USA, July 21-26, 2017}, 2017, pp. 3376--3385. [Online]. Available:
  \url{https://doi.org/10.1109/CVPR.2017.360}
\BIBentrySTDinterwordspacing

\bibitem{LiYC18}
\BIBentryALTinterwordspacing
L.~Li, H.~Yang, and L.~Chen, ``Spatial invariant person search network,'' in
  \emph{Pattern Recognition and Computer Vision - First Chinese Conference,
  {PRCV} 2018, Guangzhou, China, November 23-26, 2018, Proceedings, Part {II}},
  2018, pp. 122--133. [Online]. Available:
  \url{https://doi.org/10.1007/978-3-030-03335-4\_11}
\BIBentrySTDinterwordspacing

\bibitem{LiuFJKZQJY17}
\BIBentryALTinterwordspacing
H.~Liu, J.~Feng, Z.~Jie, J.~Karlekar, B.~Zhao, M.~Qi, J.~Jiang, and S.~Yan,
  ``Neural person search machines,'' in \emph{{IEEE} International Conference
  on Computer Vision, {ICCV} 2017, Venice, Italy, October 22-29, 2017}, 2017,
  pp. 493--501. [Online]. Available: \url{https://doi.org/10.1109/ICCV.2017.61}
\BIBentrySTDinterwordspacing

\bibitem{DBLP:conf/cvpr/MunjalATG19}
B.~Munjal, S.~Amin, F.~Tombari, and F.~Galasso, ``Query-guided end-to-end
  person search,'' in \emph{{IEEE} Conference on Computer Vision and Pattern
  Recognition, {CVPR} 2019, Long Beach, CA, USA, June 16-20, 2019}, 2019, pp.
  811--820.

\bibitem{ZhengZSCYT17}
\BIBentryALTinterwordspacing
L.~Zheng, H.~Zhang, S.~Sun, M.~Chandraker, Y.~Yang, and Q.~Tian, ``Person
  re-identification in the wild,'' in \emph{2017 {IEEE} Conference on Computer
  Vision and Pattern Recognition, {CVPR} 2017, Honolulu, HI, USA, July 21-26,
  2017}, 2017, pp. 3346--3355. [Online]. Available:
  \url{https://doi.org/10.1109/CVPR.2017.357}
\BIBentrySTDinterwordspacing

\bibitem{LanZG18}
\BIBentryALTinterwordspacing
X.~Lan, X.~Zhu, and S.~Gong, ``Person search by multi-scale matching,'' in
  \emph{Computer Vision - {ECCV} 2018 - 15th European Conference, Munich,
  Germany, September 8-14, 2018, Proceedings, Part {I}}, 2018, pp. 553--569.
  [Online]. Available: \url{https://doi.org/10.1007/978-3-030-01246-5\_33}
\BIBentrySTDinterwordspacing

\bibitem{ChenZOYT18}
\BIBentryALTinterwordspacing
D.~Chen, S.~Zhang, W.~Ouyang, J.~Yang, and Y.~Tai, ``Person search via a
  mask-guided two-stream {CNN} model,'' in \emph{Computer Vision - {ECCV} 2018
  - 15th European Conference, Munich, Germany, September 8-14, 2018,
  Proceedings, Part {VII}}, 2018, pp. 764--781. [Online]. Available:
  \url{https://doi.org/10.1007/978-3-030-01234-2\_45}
\BIBentrySTDinterwordspacing

\bibitem{DBLP:conf/cvpr/YanZNZXY19}
Y.~Yan, Q.~Zhang, B.~Ni, W.~Zhang, M.~Xu, and X.~Yang, ``Learning context graph
  for person search,'' in \emph{{IEEE} Conference on Computer Vision and
  Pattern Recognition, {CVPR} 2019, Long Beach, CA, USA, June 16-20, 2019},
  2019, pp. 2158--2167.

\bibitem{RenHG017}
\BIBentryALTinterwordspacing
S.~Ren, K.~He, R.~B. Girshick, and J.~Sun, ``Faster {R-CNN:} towards real-time
  object detection with region proposal networks,'' \emph{{IEEE} Trans. Pattern
  Anal. Mach. Intell.}, vol.~39, no.~6, pp. 1137--1149, 2017. [Online].
  Available: \url{https://doi.org/10.1109/TPAMI.2016.2577031}
\BIBentrySTDinterwordspacing

\bibitem{HeZRS15}
\BIBentryALTinterwordspacing
K.~He, X.~Zhang, S.~Ren, and J.~Sun, ``Deep residual learning for image
  recognition,'' \emph{CoRR}, vol. abs/1512.03385, 2015. [Online]. Available:
  \url{http://arxiv.org/abs/1512.03385}
\BIBentrySTDinterwordspacing

\bibitem{HuangLMW17}
\BIBentryALTinterwordspacing
G.~Huang, Z.~Liu, L.~van~der Maaten, and K.~Q. Weinberger, ``Densely connected
  convolutional networks,'' in \emph{2017 {IEEE} Conference on Computer Vision
  and Pattern Recognition, {CVPR} 2017, Honolulu, HI, USA, July 21-26, 2017},
  2017, pp. 2261--2269. [Online]. Available:
  \url{https://doi.org/10.1109/CVPR.2017.243}
\BIBentrySTDinterwordspacing

\bibitem{HuSS18}
\BIBentryALTinterwordspacing
J.~Hu, L.~Shen, and G.~Sun, ``Squeeze-and-excitation networks,'' in \emph{2018
  {IEEE} Conference on Computer Vision and Pattern Recognition, {CVPR} 2018,
  Salt Lake City, UT, USA, June 18-22, 2018}, 2018, pp. 7132--7141. [Online].
  Available:
  \url{http://openaccess.thecvf.com/content\_cvpr\_2018/html/Hu\_Squeeze-and-Excitation\_Networks\_CVPR\_2018\_paper.html}
\BIBentrySTDinterwordspacing

\bibitem{ChangHX18}
\BIBentryALTinterwordspacing
X.~Chang, T.~M. Hospedales, and T.~Xiang, ``Multi-level factorisation net for
  person re-identification,'' in \emph{2018 {IEEE} Conference on Computer
  Vision and Pattern Recognition, {CVPR} 2018, Salt Lake City, UT, USA, June
  18-22, 2018}, 2018, pp. 2109--2118. [Online]. Available:
  \url{http://openaccess.thecvf.com/content\_cvpr\_2018/html/Chang\_Multi-Level\_Factorisation\_Net\_CVPR\_2018\_paper.html}
\BIBentrySTDinterwordspacing

\bibitem{SunZYTW18}
\BIBentryALTinterwordspacing
Y.~Sun, L.~Zheng, Y.~Yang, Q.~Tian, and S.~Wang, ``Beyond part models: Person
  retrieval with refined part pooling (and {A} strong convolutional
  baseline),'' in \emph{Computer Vision - {ECCV} 2018 - 15th European
  Conference, Munich, Germany, September 8-14, 2018, Proceedings, Part {IV}},
  2018, pp. 501--518. [Online]. Available:
  \url{https://doi.org/10.1007/978-3-030-01225-0\_30}
\BIBentrySTDinterwordspacing

\bibitem{WangYCLZ18}
\BIBentryALTinterwordspacing
G.~Wang, Y.~Yuan, X.~Chen, J.~Li, and X.~Zhou, ``Learning discriminative
  features with multiple granularities for person re-identification,'' in
  \emph{2018 {ACM} Multimedia Conference on Multimedia Conference, {MM} 2018,
  Seoul, Republic of Korea, October 22-26, 2018}, 2018, pp. 274--282. [Online].
  Available: \url{https://doi.org/10.1145/3240508.3240552}
\BIBentrySTDinterwordspacing

\bibitem{ZhangWBLL18}
\BIBentryALTinterwordspacing
S.~Zhang, L.~Wen, X.~Bian, Z.~Lei, and S.~Z. Li, ``Occlusion-aware {R-CNN:}
  detecting pedestrians in a crowd,'' in \emph{Computer Vision - {ECCV} 2018 -
  15th European Conference, Munich, Germany, September 8-14, 2018, Proceedings,
  Part {III}}, 2018, pp. 657--674. [Online]. Available:
  \url{https://doi.org/10.1007/978-3-030-01219-9\_39}
\BIBentrySTDinterwordspacing

\bibitem{NohLKK18}
J.~Noh, S.~Lee, B.~Kim, and G.~Kim, ``Improving occlusion and hard negative
  handling for single-stage pedestrian detectors,'' in \emph{2018 {IEEE}
  Conference on Computer Vision and Pattern Recognition, {CVPR} 2018, Salt Lake
  City, UT, USA, June 18-22, 2018}, 2018, pp. 966--974.

\bibitem{SongSXSP18}
\BIBentryALTinterwordspacing
T.~Song, L.~Sun, D.~Xie, H.~Sun, and S.~Pu, ``Small-scale pedestrian detection
  based on topological line localization and temporal feature aggregation,'' in
  \emph{Computer Vision - {ECCV} 2018 - 15th European Conference, Munich,
  Germany, September 8-14, 2018, Proceedings, Part {VII}}, 2018, pp. 554--569.
  [Online]. Available: \url{https://doi.org/10.1007/978-3-030-01234-2\_33}
\BIBentrySTDinterwordspacing

\bibitem{LiLSXFY18}
\BIBentryALTinterwordspacing
J.~Li, X.~Liang, S.~Shen, T.~Xu, J.~Feng, and S.~Yan, ``Scale-aware fast
  {R-CNN} for pedestrian detection,'' \emph{{IEEE} Trans. Multimedia}, vol.~20,
  no.~4, pp. 985--996, 2018. [Online]. Available:
  \url{https://doi.org/10.1109/TMM.2017.2759508}
\BIBentrySTDinterwordspacing

\bibitem{OuyangW13}
\BIBentryALTinterwordspacing
W.~Ouyang and X.~Wang, ``Joint deep learning for pedestrian detection,'' in
  \emph{{IEEE} International Conference on Computer Vision, {ICCV} 2013,
  Sydney, Australia, December 1-8, 2013}, 2013, pp. 2056--2063. [Online].
  Available: \url{https://doi.org/10.1109/ICCV.2013.257}
\BIBentrySTDinterwordspacing

\bibitem{ZhouY18}
\BIBentryALTinterwordspacing
C.~Zhou and J.~Yuan, ``Bi-box regression for pedestrian detection and occlusion
  estimation,'' in \emph{Computer Vision - {ECCV} 2018 - 15th European
  Conference, Munich, Germany, September 8-14, 2018, Proceedings, Part {I}},
  2018, pp. 138--154. [Online]. Available:
  \url{https://doi.org/10.1007/978-3-030-01246-5\_9}
\BIBentrySTDinterwordspacing

\bibitem{DBLP:journals/tmm/WangCLWZ18}
\BIBentryALTinterwordspacing
S.~Wang, J.~Cheng, H.~Liu, F.~Wang, and H.~Zhou, ``Pedestrian detection via
  body part semantic and contextual information with {DNN},'' \emph{{IEEE}
  Trans. Multimedia}, vol.~20, no.~11, pp. 3148--3159, 2018. [Online].
  Available: \url{https://doi.org/10.1109/TMM.2018.2829602}
\BIBentrySTDinterwordspacing

\bibitem{FelzenszwalbGMR10}
\BIBentryALTinterwordspacing
P.~F. Felzenszwalb, R.~B. Girshick, D.~A. McAllester, and D.~Ramanan, ``Object
  detection with discriminatively trained part-based models,'' \emph{{IEEE}
  Trans. Pattern Anal. Mach. Intell.}, vol.~32, no.~9, pp. 1627--1645, 2010.
  [Online]. Available: \url{https://doi.org/10.1109/TPAMI.2009.167}
\BIBentrySTDinterwordspacing

\bibitem{GirshickDDM14}
\BIBentryALTinterwordspacing
R.~B. Girshick, J.~Donahue, T.~Darrell, and J.~Malik, ``Rich feature
  hierarchies for accurate object detection and semantic segmentation,'' in
  \emph{2014 {IEEE} Conference on Computer Vision and Pattern Recognition,
  {CVPR} 2014, Columbus, OH, USA, June 23-28, 2014}, 2014, pp. 580--587.
  [Online]. Available: \url{https://doi.org/10.1109/CVPR.2014.81}
\BIBentrySTDinterwordspacing

\bibitem{Girshick15}
\BIBentryALTinterwordspacing
R.~B. Girshick, ``Fast {R-CNN},'' in \emph{2015 {IEEE} International Conference
  on Computer Vision, {ICCV} 2015, Santiago, Chile, December 7-13, 2015}, 2015,
  pp. 1440--1448. [Online]. Available:
  \url{https://doi.org/10.1109/ICCV.2015.169}
\BIBentrySTDinterwordspacing

\bibitem{YiLLL14}
\BIBentryALTinterwordspacing
D.~Yi, Z.~Lei, S.~Liao, and S.~Z. Li, ``Deep metric learning for person
  re-identification,'' in \emph{22nd International Conference on Pattern
  Recognition, {ICPR} 2014, Stockholm, Sweden, August 24-28, 2014}, 2014, pp.
  34--39. [Online]. Available: \url{https://doi.org/10.1109/ICPR.2014.16}
\BIBentrySTDinterwordspacing

\bibitem{LiZXW14}
\BIBentryALTinterwordspacing
W.~Li, R.~Zhao, T.~Xiao, and X.~Wang, ``Deepreid: Deep filter pairing neural
  network for person re-identification,'' in \emph{2014 {IEEE} Conference on
  Computer Vision and Pattern Recognition, {CVPR} 2014, Columbus, OH, USA, June
  23-28, 2014}, 2014, pp. 152--159. [Online]. Available:
  \url{https://doi.org/10.1109/CVPR.2014.27}
\BIBentrySTDinterwordspacing

\bibitem{AhmedJM15}
\BIBentryALTinterwordspacing
E.~Ahmed, M.~J. Jones, and T.~K. Marks, ``An improved deep learning
  architecture for person re-identification,'' in \emph{{IEEE} Conference on
  Computer Vision and Pattern Recognition, {CVPR} 2015, Boston, MA, USA, June
  7-12, 2015}, 2015, pp. 3908--3916. [Online]. Available:
  \url{https://doi.org/10.1109/CVPR.2015.7299016}
\BIBentrySTDinterwordspacing

\bibitem{VariorSLXW16}
\BIBentryALTinterwordspacing
R.~R. Varior, B.~Shuai, J.~Lu, D.~Xu, and G.~Wang, ``A siamese long short-term
  memory architecture for human re-identification,'' in \emph{Computer Vision -
  {ECCV} 2016 - 14th European Conference, Amsterdam, The Netherlands, October
  11-14, 2016, Proceedings, Part {VII}}, 2016, pp. 135--153. [Online].
  Available: \url{https://doi.org/10.1007/978-3-319-46478-7\_9}
\BIBentrySTDinterwordspacing

\bibitem{ChenZW16}
J.~Chen, Z.~Zhang, and Y.~Wang, ``Corrections to "relevance metric learning for
  person re-identification by exploiting listwise similarities",'' \emph{{IEEE}
  Trans. Image Processing}, vol.~25, no.~1, p. 494, 2016.

\bibitem{DingLWC15}
\BIBentryALTinterwordspacing
S.~Ding, L.~Lin, G.~Wang, and H.~Chao, ``Deep feature learning with relative
  distance comparison for person re-identification,'' \emph{Pattern
  Recognition}, vol.~48, no.~10, pp. 2993--3003, 2015. [Online]. Available:
  \url{https://doi.org/10.1016/j.patcog.2015.04.005}
\BIBentrySTDinterwordspacing

\bibitem{ChengGZWZ16}
\BIBentryALTinterwordspacing
D.~Cheng, Y.~Gong, S.~Zhou, J.~Wang, and N.~Zheng, ``Person re-identification
  by multi-channel parts-based {CNN} with improved triplet loss function,'' in
  \emph{2016 {IEEE} Conference on Computer Vision and Pattern Recognition,
  {CVPR} 2016, Las Vegas, NV, USA, June 27-30, 2016}, 2016, pp. 1335--1344.
  [Online]. Available: \url{https://doi.org/10.1109/CVPR.2016.149}
\BIBentrySTDinterwordspacing

\bibitem{WangZLZZ16}
\BIBentryALTinterwordspacing
F.~Wang, W.~Zuo, L.~Lin, D.~Zhang, and L.~Zhang, ``Joint learning of
  single-image and cross-image representations for person re-identification,''
  in \emph{2016 {IEEE} Conference on Computer Vision and Pattern Recognition,
  {CVPR} 2016, Las Vegas, NV, USA, June 27-30, 2016}, 2016, pp. 1288--1296.
  [Online]. Available: \url{https://doi.org/10.1109/CVPR.2016.144}
\BIBentrySTDinterwordspacing

\bibitem{DBLP:journals/tmm/ZhouWSHGZ18}
\BIBentryALTinterwordspacing
S.~Zhou, J.~Wang, R.~Shi, Q.~Hou, Y.~Gong, and N.~Zheng, ``Large margin
  learning in set-to-set similarity comparison for person reidentification,''
  \emph{{IEEE} Trans. Multimedia}, vol.~20, no.~3, pp. 593--604, 2018.
  [Online]. Available: \url{https://doi.org/10.1109/TMM.2017.2755983}
\BIBentrySTDinterwordspacing

\bibitem{LiC0H17}
\BIBentryALTinterwordspacing
D.~Li, X.~Chen, Z.~Zhang, and K.~Huang, ``Learning deep context-aware features
  over body and latent parts for person re-identification,'' in \emph{2017
  {IEEE} Conference on Computer Vision and Pattern Recognition, {CVPR} 2017,
  Honolulu, HI, USA, July 21-26, 2017}, 2017, pp. 7398--7407. [Online].
  Available: \url{https://doi.org/10.1109/CVPR.2017.782}
\BIBentrySTDinterwordspacing

\bibitem{YaoZHZX019}
\BIBentryALTinterwordspacing
H.~Yao, S.~Zhang, R.~Hong, Y.~Zhang, C.~Xu, and Q.~Tian, ``Deep representation
  learning with part loss for person re-identification,'' \emph{{IEEE} Trans.
  Image Processing}, vol.~28, no.~6, pp. 2860--2871, 2019. [Online]. Available:
  \url{https://doi.org/10.1109/TIP.2019.2891888}
\BIBentrySTDinterwordspacing

\bibitem{ZhaoLZW17}
\BIBentryALTinterwordspacing
L.~Zhao, X.~Li, Y.~Zhuang, and J.~Wang, ``Deeply-learned part-aligned
  representations for person re-identification,'' in \emph{{IEEE} International
  Conference on Computer Vision, {ICCV} 2017, Venice, Italy, October 22-29,
  2017}, 2017, pp. 3239--3248. [Online]. Available:
  \url{https://doi.org/10.1109/ICCV.2017.349}
\BIBentrySTDinterwordspacing

\bibitem{ZhengHLY19}
L.~Zheng, Y.~Huang, H.~Lu, and Y.~Yang, ``Pose-invariant embedding for deep
  person re-identification,'' \emph{{IEEE} Trans. Image Processing}, vol.~28,
  no.~9, pp. 4500--4509, 2019.

\bibitem{ZhaoTSSYYWT17}
\BIBentryALTinterwordspacing
H.~Zhao, M.~Tian, S.~Sun, J.~Shao, J.~Yan, S.~Yi, X.~Wang, and X.~Tang,
  ``Spindle net: Person re-identification with human body region guided feature
  decomposition and fusion,'' in \emph{2017 {IEEE} Conference on Computer
  Vision and Pattern Recognition, {CVPR} 2017, Honolulu, HI, USA, July 21-26,
  2017}, 2017, pp. 907--915. [Online]. Available:
  \url{https://doi.org/10.1109/CVPR.2017.103}
\BIBentrySTDinterwordspacing

\bibitem{SuLZX0T17}
\BIBentryALTinterwordspacing
C.~Su, J.~Li, S.~Zhang, J.~Xing, W.~Gao, and Q.~Tian, ``Pose-driven deep
  convolutional model for person re-identification,'' in \emph{{IEEE}
  International Conference on Computer Vision, {ICCV} 2017, Venice, Italy,
  October 22-29, 2017}, 2017, pp. 3980--3989. [Online]. Available:
  \url{https://doi.org/10.1109/ICCV.2017.427}
\BIBentrySTDinterwordspacing

\bibitem{DBLP:journals/tmm/WeiZY0019}
\BIBentryALTinterwordspacing
L.~Wei, S.~Zhang, H.~Yao, W.~Gao, and Q.~Tian, ``{GLAD:} global-local-alignment
  descriptor for scalable person re-identification,'' \emph{{IEEE} Trans.
  Multimedia}, vol.~21, no.~4, pp. 986--999, 2019. [Online]. Available:
  \url{https://doi.org/10.1109/TMM.2018.2870522}
\BIBentrySTDinterwordspacing

\bibitem{ZhouWMLGZ19}
S.~Zhou, J.~Wang, D.~Meng, Y.~Liang, Y.~Gong, and N.~Zheng, ``Discriminative
  feature learning with foreground attention for person re-identification,''
  \emph{{IEEE} Trans. Image Processing}, vol.~28, no.~9, pp. 4671--4684, 2019.

\bibitem{LiuFQJY17}
\BIBentryALTinterwordspacing
H.~Liu, J.~Feng, M.~Qi, J.~Jiang, and S.~Yan, ``End-to-end comparative
  attention networks for person re-identification,'' \emph{{IEEE} Trans. Image
  Processing}, vol.~26, no.~7, pp. 3492--3506, 2017. [Online]. Available:
  \url{https://doi.org/10.1109/TIP.2017.2700762}
\BIBentrySTDinterwordspacing

\bibitem{LiZG18}
W.~Li, X.~Zhu, and S.~Gong, ``Harmonious attention network for person
  re-identification,'' in \emph{2018 {IEEE} Conference on Computer Vision and
  Pattern Recognition, {CVPR} 2018, Salt Lake City, UT, USA, June 18-22, 2018},
  2018, pp. 2285--2294.

\bibitem{LiuZTSSYYW17}
\BIBentryALTinterwordspacing
X.~Liu, H.~Zhao, M.~Tian, L.~Sheng, J.~Shao, S.~Yi, J.~Yan, and X.~Wang,
  ``Hydraplus-net: Attentive deep features for pedestrian analysis,'' in
  \emph{{IEEE} International Conference on Computer Vision, {ICCV} 2017,
  Venice, Italy, October 22-29, 2017}, 2017, pp. 350--359. [Online]. Available:
  \url{https://doi.org/10.1109/ICCV.2017.46}
\BIBentrySTDinterwordspacing

\bibitem{ZhaoOW13}
\BIBentryALTinterwordspacing
R.~Zhao, W.~Ouyang, and X.~Wang, ``Unsupervised salience learning for person
  re-identification,'' in \emph{2013 {IEEE} Conference on Computer Vision and
  Pattern Recognition, Portland, OR, USA, June 23-28, 2013}, 2013, pp.
  3586--3593. [Online]. Available: \url{https://doi.org/10.1109/CVPR.2013.460}
\BIBentrySTDinterwordspacing

\bibitem{KostingerHWRB12}
\BIBentryALTinterwordspacing
M.~K{\"{o}}stinger, M.~Hirzer, P.~Wohlhart, P.~M. Roth, and H.~Bischof, ``Large
  scale metric learning from equivalence constraints,'' in \emph{2012 {IEEE}
  Conference on Computer Vision and Pattern Recognition, Providence, RI, USA,
  June 16-21, 2012}, 2012, pp. 2288--2295. [Online]. Available:
  \url{https://doi.org/10.1109/CVPR.2012.6247939}
\BIBentrySTDinterwordspacing

\bibitem{han2019re}
C.~Han, J.~Ye, Y.~Zhong, X.~Tan, C.~Zhang, C.~Gao, and N.~Sang, ``Re-id driven
  localization refinement for person search,'' in \emph{Proceedings of the IEEE
  International Conference on Computer Vision}, 2019, pp. 9814--9823.

\end{thebibliography}
%

%

\end{document}